\def\TIPcleanminorrevision{1}
\documentclass[lettersize,journal]{IEEEtran}
\usepackage{amsmath, amsfonts, amssymb, amsthm, mathtools}
\usepackage{array}
\newcolumntype{C}[1]{>{\centering\arraybackslash}p{#1}}
\usepackage[caption=false,font=scriptsize,labelfont=sf,textfont=sf]{subfig}  
\usepackage{url}
\usepackage{graphicx}

\usepackage{xspace}
\usepackage{booktabs}
\usepackage{pifont}
\usepackage{cite}
\usepackage[dvipsnames, table]{xcolor}
\usepackage[normalem]{ulem}
\usepackage{soul}
\usepackage{multirow}
\newtheorem{theorem}{Theorem}
\newtheorem{lemma}{Lemma}

\newif\ifshowrev
\showrevfalse

\DeclareRobustCommand{\rev}[1]{%
  \ifshowrev\textcolor{blue}{#1}\else#1\fi
}

\newenvironment{revblock}{%
  \begingroup
  \ifshowrev
    \color{blue}%
    \arrayrulecolor{blue}%
  \fi
}{%
  \ifshowrev
    \arrayrulecolor{black}%
  \fi
  \endgroup
}

\newif\ifshowminorrev
\ifdefined\TIPcleanminorrevision
  \showminorrevfalse
\else
  \showminorrevtrue
\fi

\colorlet{minorrevcolor}{blue}
\colorlet{minorrevdeletedtext}{black}

\DeclareRobustCommand{\minorrev}[1]{%
  \ifshowminorrev
    \textcolor{minorrevcolor}{#1}%
  \else
    #1%
  \fi
}

\DeclareRobustCommand{\minorrevstrike}{%
  \bgroup
  \color{minorrevdeletedtext}%
  \markoverwith{%
    \textcolor{minorrevcolor}{\rule[0.5ex]{2pt}{0.4pt}}%
  }%
  \ULon
}

\DeclareRobustCommand{\minorrevdel}[1]{%
  \ifshowminorrev
    \minorrevstrike{#1}%
  \else
    \unskip
  \fi
}

\DeclareRobustCommand{\minorrevreplace}[2]{%
  \ifshowminorrev
    \minorrevstrike{#1}\space
    \textcolor{minorrevcolor}{#2}%
  \else
    #2%
  \fi
}

\newenvironment{minorrevblock}{%
  \begingroup
  \ifshowminorrev
    \color{minorrevcolor}%
    \arrayrulecolor{minorrevcolor}%
  \fi
}{%
  \ifshowminorrev
    \arrayrulecolor{black}%
  \fi
  \endgroup
}

\newcommand{\ie}{\textit{i.e.},\xspace}
\newcommand{\eg}{\textit{e.g.},\xspace}
\newcommand{\etal}{\textit{et al.}\xspace}

\newcommand{\ghl}[1]{%
  \begingroup
  \sethlcolor{ForestGreen!25}%
  \hl{#1}%
  \endgroup
}

\begin{document}
\title{ZIP: Scalable Crowd Counting via Zero‑Inflated Poisson Modeling}
\author{
Yiming~Ma,~\IEEEmembership{Member,~IEEE,}
Victor~Sanchez,~\IEEEmembership{Member,~IEEE,}
and~Tanaya~Guha,~\IEEEmembership{Senior Member,~IEEE}
\thanks{\textcopyright\ 2026 IEEE. Personal use of this material is permitted. Permission from IEEE must be obtained for all other uses, in any current or future media, including reprinting/republishing this material for advertising or promotional purposes, creating new collective works, for resale or redistribution to servers or lists, or reuse of any copyrighted component of this work in other works.}

\thanks{This is the author accepted manuscript of the article published in IEEE Transactions on Image Processing, vol. 35, pp. 11982--11997, 2026. DOI: 10.1109/TIP.2026.3752245.}

\thanks{This work was supported in part by the Spanish Ministry of Science, Innovation and Universities, Spain; and in part by the Agencia Estatal de Investigaci\'on--ATRAE Program under Grant ATR2024-154513. (Corresponding author: Yiming Ma.)}

\thanks{Yiming Ma was with the Department of Computer Science, University of Warwick, Coventry CV4 7AL, U.K. He is now with the School of Biomedical Engineering and Imaging Sciences, King's College London, London, U.K. (e-mail: yiming.ma.1@warwick.ac.uk; yiming.2.ma@kcl.ac.uk).}

\thanks{Victor Sanchez is with the Department of Computer Science, University of Warwick, Coventry CV4 7AL, U.K., and is also with Universitat Aut\`onoma de Barcelona, Barcelona, Spain (e-mail: v.f.sanchez-silva@warwick.ac.uk).}

\thanks{Tanaya Guha is with the School of Computing Science, University of Glasgow, Glasgow G12 8RZ, U.K. (e-mail: tanaya.guha@glasgow.ac.uk).}
}

\maketitle

\begin{abstract}
The majority of crowd counting methods directly regress blockwise density maps using losses based on Mean Squared Error (MSE).
This practice has two critical limitations:
1) it fails to account for the extreme spatial sparsity of annotations (\ie over 95\% of standard 8$\times$8 blocks are empty across most benchmarks), which makes supervision signals in informative regions diluted by the predominant zeros. 
2) MSE corresponds to a Gaussian error model that poorly matches discrete, non‑negative count data.
We introduce a novel crowd counting framework that is scalable and directly addresses the above limitations. This framework, named \emph{ZIP}, models blockwise counts with a \emph{Zero‑Inflated Poisson} likelihood: 
A zero-inflation term models excess zeros in sparse annotations, while the Poisson component captures expected counts while respecting their discrete, non-negative nature.
%
We provide a generalization analysis showing that the ZIP negative log-likelihood (NLL) has the same order of uniform generalization gap as Poisson NLL while offering a richer zero-inflated likelihood family, and \minorrevreplace{
scales more favorably than OT-based supervision with respect to the
number of spatial blocks
}{
that its bound scales more favorably with the number of spatial blocks
than optimal-transport-based supervision}.
To assess the scalability and broad applicability of ZIP, we apply the same formulation across diverse backbones, from lightweight models to high-capacity architectures, spanning more than 100$\times$ in both parameter count and inference FLOPs.
Experiments on four standard crowd-counting benchmarks show that ZIP achieves state-of-the-art or highly competitive performance across a wide range of backbone capacities.
%
Although ZIP is designed for sparse blockwise count maps, it also remains effective in dense crowd scenes, reducing MAE and RMSE by 3.9\% and 8.5\%, respectively, over mPrompt on UCF-QNRF.
Code and trained weights are available at \url{https://github.com/Yiming-M/ZIP}.
\end{abstract}


\begin{figure}[!t]
\centering
\subfloat[Input Image\label{fig:vis_image}]{
    \includegraphics[width=0.47\linewidth]{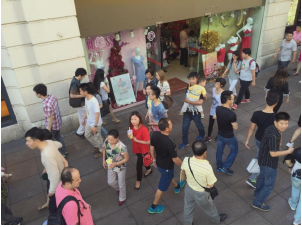}
}%
\hfil
\subfloat[Ground Truth Density Map\label{fig:vis_gt_den}]{
    \includegraphics[width=0.47\linewidth]{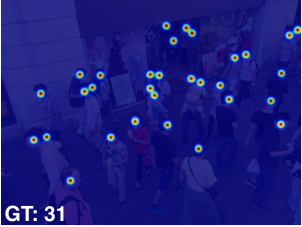}
}%
\vfil
\subfloat[Structural Zero Map\label{fig:vis_zero}]{
    \includegraphics[width=0.47\linewidth]{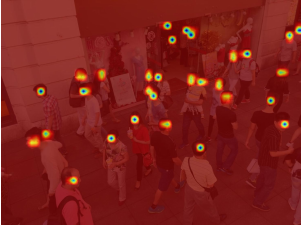}
}%
\hfil
\subfloat[Predicted Density Map\label{fig:vis_pred_den}]{
    \includegraphics[width=0.47\linewidth]{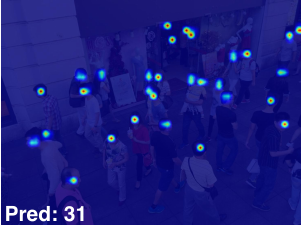}
}%
\caption{
\minorrevreplace{The structural zero map output by ZIP assigns high probability to non-head-central regions and highlights candidate head-center areas.}{The learned zero-inflation map $\boldsymbol{\pi}$ assigns high values to non-head-central regions, while its complement $(1-\boldsymbol{\pi})$ highlights candidate head-center areas.}
This visualization uses an image from the ShanghaiTech B \cite{zhang2016single} test split.
\textbf{Color scale (all panels):} red = higher value, blue = lower value (per-panel normalization).
In (c),
red regions have a \emph{high} probability of being structurally empty (background, torso, peripheral head parts), blue indicates \emph{low} structural-zero probability (\ie candidate head centers).
In (d),
red regions correspond to a \emph{high} expected local crowd density near head centers; blue denotes near-zero density. 
This structural-zero modeling addresses density map sparsity by softly down-weighting semantically empty regions.
}
\label{fig:vis}
\end{figure}

\section{Introduction}
\IEEEPARstart{C}{rowd} counting aims to estimate the number of people present in an image or a video frame. It has a broad range of real-world applications, including public-space analytics \cite{valencia2021vision} 
and intelligent transportation systems \cite{mccarthy2025video}.
The majority of existing crowd counting methods \cite{ma2019bayesian, cheng2022rethinking, han2023steerer} construct blockwise training targets by first convolving point annotations with Gaussian kernels to create a pixel-level density map, and then summing density values within each $8\times8$ block to obtain blockwise density maps. Models are trained to regress these blockwise targets (usually with Mean Squared Error (MSE)) in lieu of raw pixel supervision to mitigate extreme pixel-level sparsity.

In the above methods, however, the aggregated blockwise density maps remain \textbf{highly sparse}: across ShanghaiTech A \& B~\cite{zhang2016single}, UCF-QNRF~\cite{idrees2018composition}, and NWPU-Crowd~\cite{wang2020nwpu}, over 95\% of $8\times8$ blocks contain zero people (empirical measurement). This zero dominance skews squared-error losses, dilutes supervision from populated regions, and biases models toward under-counting.
These methods also suffer from \textbf{label noise} introduced by Gaussian smoothing. Point annotations are spatially uncertain: a human click marks an approximate head center, and nearby pixels could be equally valid. Spreading each point with a local kernel is intended to make supervision tolerant to such subjective variation. However, since the true head extents are unknown, mismatched kernel bandwidths yield over- or under-counts after block aggregation and thus inject noise into learning targets.

Prior work made attempts to mitigate sparsity by first segmenting foreground crowd regions and then counting only within those areas~\cite{rong2021coarse,modolo2021understanding,guo2024regressor}. Such approaches introduce additional multi-task loss balancing complexity. Also, since crowd counting datasets are not annotated with segmentation masks, these approaches usually derive pseudo masks from Gaussian-smoothed density maps, inheriting the same head-size uncertainty and label noise discussed above.

\begin{figure}[t]
    \centering
    \includegraphics[width=1.0\linewidth]{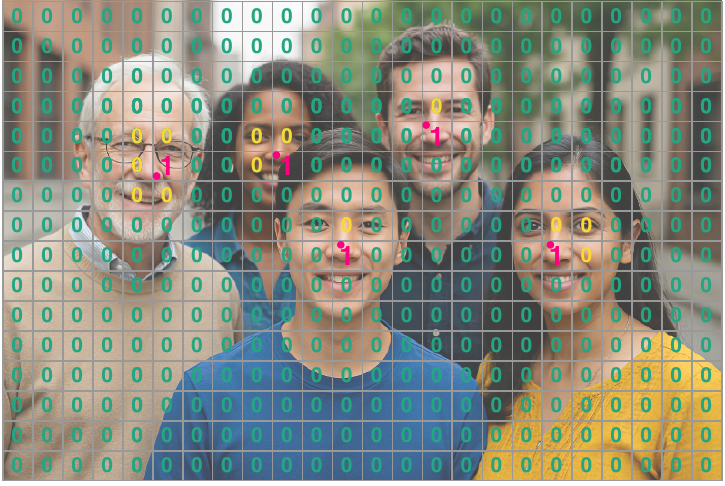}
    \caption{Illustration of the concept of structural and sampling zeros in ZIP (best viewed in color).
    The image (synthesized for illustration purposes) is overlaid with a grid where each cell represents a spatial block.
    The \textcolor[HTML]{FF0080}{red} dot marks a ground-truth head annotation, and the block containing it is labeled with the \textcolor[HTML]{FF0080}{red} number ``1''.
    \textcolor[HTML]{FDD835}{Yellow} zeros, which indicate sampling zeros, are assigned to blocks that correspond to the head-center region but receive zero count due to the point-based annotation protocol.
    \textcolor[HTML]{24A683}{Green} zeros denote structural zeros, corresponding to background regions, non-head body parts or outer regions of the head that are not associated with any annotations.}
    \label{fig:zeros}
\end{figure}

In addition, MSE-based loss functions present a \textbf{modeling mismatch}. Blockwise targets represent discrete, non-negative, and zero-heavy count data, whereas MSE implicitly assumes Gaussian residuals on continuous values. A Poisson model can capture both the discreteness and non-negativity of counts, yet a standard Poisson struggles with the overwhelming presence of zeros.
These gaps motivate a formulation that 
\minorrevreplace{
explicitly separates structural emptiness from stochastic counts
}{
models excess zeros and stochastic counts with separate components
}, leading to our Zero-Inflated Poisson (ZIP) framework.

\minorrevreplace{
We model zeros in blockwise density maps as arising from two distinct mechanisms:
}{
Within the ZIP formulation, an observed zero count can arise through two latent mechanisms:
}
\begin{itemize}
    \item \textbf{Structural zeros}, which are blocks that are deterministically zero due to their semantic irrelevance. These include background (sky, buildings, etc.), body parts (limbs, chest, etc.) and peripheral head regions that do not correspond to head centers\minorrevdel{, and they comprise the majority of zero blocks}. Fig.~\ref{fig:vis_zero} \minorrevreplace{illustrates this type of zeros predicted by ZIP}{visualizes the learned zero-inflation map} on ShanghaiTech B.
    \item \textbf{Sampling zeros}: each head is annotated by a \emph{single point} which is assigned to a unique supervision block. Neighboring blocks that also correspond to the head center therefore receive zero count. 
    Modeling these annotation effects in the Poisson component absorbs the positional ambiguity that Gaussian smoothing is meant to solve, removing kernel bandwidth tuning.
\end{itemize}
Fig.~\ref{fig:zeros} provides a schematic comparison of these two types of zeros.
\minorrevreplace{
Because structural and sampling zeros are not annotated separately, our ZIP model learns to distinguish these two types of zeros implicitly by optimizing the negative log-likelihood of the ZIP distribution, with the Poisson rate additionally supervised by a cross entropy loss to stabilize training.
}{
Because the source of an observed zero is latent, the zero-inflation and Poisson components are learned jointly, rather than supervised separately.
}
Empirically, 
\minorrevreplace{
learned zero-inflation correlates with non-head-center areas
}{
the learned $\boldsymbol{\pi}$ map exhibits a clear spatial pattern
}.
As illustrated in Fig.~\ref{fig:vis_zero},
\minorrevreplace{
the structural-zero map assigns high structural-zero probability to most non-head regions, while preserving only the compact blue blobs centered on heads
}{
it takes high values over most non-head regions and low values around compact head-center regions
}.
Compared with segmentation-based methods, our ZIP-based method offers several notable advantages:
1) it does not rely on segmentation masks, thereby avoiding noise introduced by Gaussian smoothing;
2) it captures structural sparsity without assuming the presence of clear object boundaries, making it better suited for highly occluded and densely crowded scenes; and
%
\rev{3) it avoids explicit segmentation supervision and the associated counting-versus-segmentation loss balancing.}

\begin{figure}[t]
    \centering
    \includegraphics[width=1.0\linewidth]{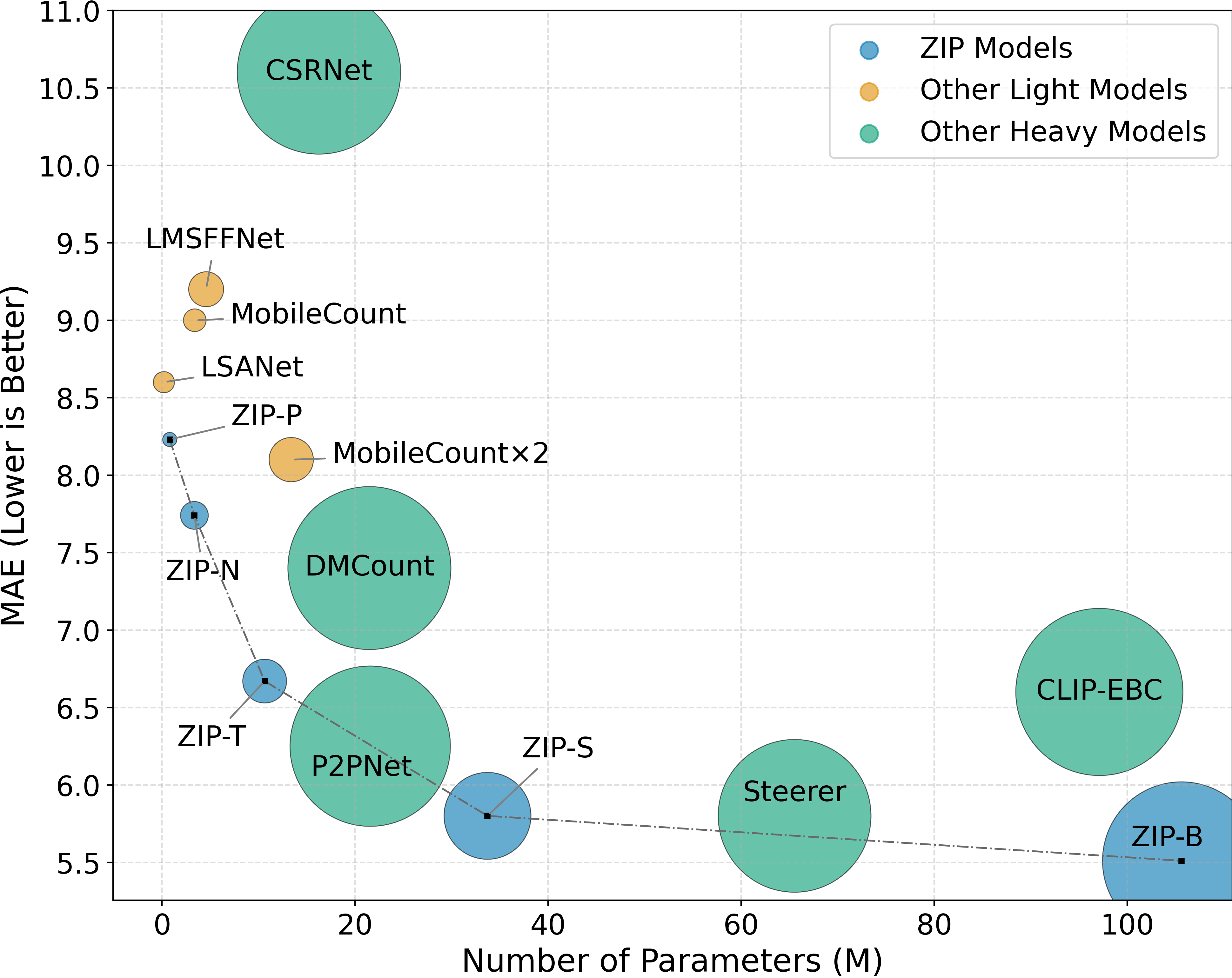}
    \caption{Scalability of the proposed ZIP framework on ShanghaiTech B.
    Each circle represents a specific model, where the radius corresponds to the number of floating point operations (FLOPs) required to process a single $1920 \times 1080$ image during inference. ZIP models (in \textcolor[HTML]{7CA9CE}{blue}) exhibit a favorable trade-off between model size and accuracy, demonstrating better scalability compared to other lightweight (\textcolor[HTML]{E0BD71}{yellow}) and heavyweight (\textcolor[HTML]{87C0AB}{green}) models.}
    \label{fig:scalability}
\end{figure}

\minorrevreplace{
Through generalization analysis and controlled experiments, we show that ZIP NLL preserves the same-order uniform generalization gap as Poisson NLL while providing a richer zero-inflated likelihood family, and scales more favorably than OT-based supervision with respect to
the number of spatial blocks.
}{
Our generalization analysis shows that a) the ZIP negative log-likelihood (NLL) preserves the same-order uniform generalization gap as Poisson NLL while providing a richer zero-inflated likelihood family, 
and b) that its worst-case bound has a lower-order dependence on the number of spatial blocks than the OT-based counterpart.
}
~\cite{wang2020distribution}
To further assess the scalability of ZIP, we evaluate its performance with backbones of varying computational complexities, including MobileNet, MobileCLIP, and CLIP-ConvNeXt.
As illustrated in Fig.~\ref{fig:scalability}, our ZIP framework consistently outperforms existing models in both accuracy (measured by Mean Absolute Error (MAE))
and computational efficiency (measured by the number of floating-point operations (FLOPs) needed to process a $1920 \times 1080$ image) across different parameter scales, demonstrating its superior scalability.
To summarize, our contributions are:
\begin{itemize}
    \item \textbf{Novel framework:} \textbf{ZIP} is a novel crowd counting framework based on \textbf{Zero-Inflated Poisson} regression. It explicitly \minorrevreplace{models the extreme sparsity of ground-truth count distributions by disentangling structural zeros from sampling uncertainty.}{models zero-heavy blockwise count distributions using a zero-inflation component for excess zeros and a Poisson component for discrete counts.}
    \item \textbf{Theoretical bound and experimental evidence:} We present theoretical risk bound analysis and thorough experiments on four crowd counting benchmark datasets. Results show that our base model, \textbf{ZIP-Base}, consistently outperforms state-of-the-art methods across four datasets.
    \item \textbf{Scalability:} A systematic study of framework-level scalability shows that ZIP generalizes well across a wide range of model sizes and architectures -- from lightweight convolutional neural networks (CNNs) to vision-language models, achieving state-of-the-art performance under varying computational constraints.
\end{itemize}

\begin{figure*}[!t]
    \centering
    \includegraphics[width=\linewidth]{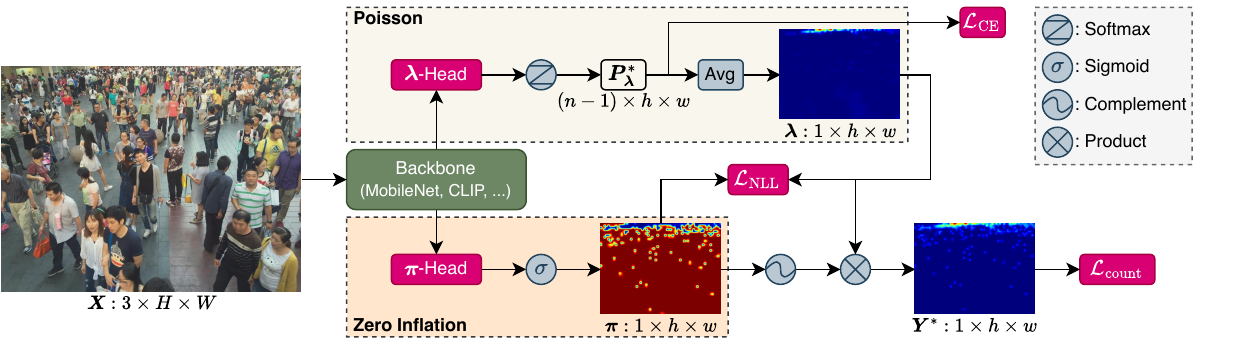}
    \caption{
    Overview of the proposed ZIP framework. 
    Given an input image $\boldsymbol{X}$ of spatial size $H \times W$, a backbone network extracts feature maps that are shared by two parallel branches: A Poisson branch (top) and a Zero Inflation branch (bottom).
    The Poisson branch (top) processes the features through a $\boldsymbol{\lambda}$-head with a softmax activation, producing the blockwise count distribution $\boldsymbol{P}^*_{\boldsymbol{\lambda}}$ over the $(n-1)$ positive bins.
    The Poisson rate per block is computed as a weighted average of bin centers, yielding the final $\boldsymbol{\lambda}$ map.
    In the Zero Inflation branch, the same features are fed to a $\boldsymbol{\pi}$-head with a sigmoid activation to estimate \minorrevreplace{the structural zero probability map}{the zero-inflation probability map} $\boldsymbol{\pi}$ of spatial size $h \times w$.
    The predicted density map $\boldsymbol{Y}^*$ is defined as the expected value of the zero-inflated Poisson distribution: $\boldsymbol{Y}^* = (1 - \boldsymbol{\pi}) \otimes \boldsymbol{\lambda}$, where $\otimes$ denotes element-wise multiplication.
    }
    \label{fig:model}
\end{figure*}

\section{Related Work}

\subsection{Handling Sparsity}

Many existing methods \cite{zhang2016single, liu2019context, ma2019bayesian, wang2020mobilecount, wang2021uniformity, han2023steerer, guo2024regressor} adopt Gaussian smoothing to address the extreme sparsity of ground-truth pixel-wise indicator density maps, where 1's correspond to labeled head centers while 0's otherwise.
These approaches typically preprocess the ground-truth density map by convolving it with Gaussian kernels, either of fixed or adaptive size. 
In principle, the kernel size should reflect the actual head size in the image; however, such information is typically unavailable in most crowd counting datasets.
As a result, Gaussian smoothing inevitably introduces noise into the ground-truth density maps, which can degrade the model's generalization ability \cite{wang2020distribution}.

There have also been some efforts \cite{rong2021coarse, modolo2021understanding, guo2024regressor} to address the sparsity by training a segmentation model which aims to separate regions containing people from the background.
However, most crowd counting datasets do not provide segmentation masks, they have to rely on Gaussian-smoothed ground-truth density maps to generate pseudo ground-truth segmentation masks.
This again introduces noise into the pseudo segmentation masks and may degrade generalization performance.
Guo \etal \cite{guo2024regressor} exploited the box annotations provided by NWPU-Crowd \cite{wang2020nwpu} to handle this problem, but compared with point annotations, box annotations are much more expensive to acquire.
Besides, these methods introduce an extra task to counting, and balancing the loss terms in such a multi-task setting poses a new challenge.

\subsection{Modeling Count Distributions in Density Maps}

On the other hand, most density-based crowd counting methods typically frame the task as a regression problem, where the model is trained to minimize the blockwise MSE between the predicted and ground-truth blockwise density maps \cite{zhang2016single, liu2019context, liu2019counting, ma2019bayesian, wang2020mobilecount, han2023steerer, guo2024regressor}.
This loss function implicitly assumes that each blockwise count follows a Gaussian distribution centered at the predicted value.
However, this formulation fails to capture the discreteness and non-negativity of count data.
DMCount \cite{wang2020distribution} addresses a different issue by reformulating density estimation as an Earth Mover's Distance (EMD) problem.
By minimizing the Wasserstein distance between predicted and ground-truth density maps, DMCount explicitly models the global transportation cost needed to match the predicted mass to the annotated points. 
This approach improves robustness to spatial misalignment and annotation noise. 
However, similar to MSE-based methods, it still operates entirely in a continuous density space, treating the density maps as divisible mass rather than integer-valued counts. 
Also, the exact computation of EMD scales poorly with the number of bins ($\mathcal{O}(n^3)$).

By comparison, Bayesian Poisson Regression (BPR) \cite{chan2009bayesian} is an early attempt to model count from a discrete perspective. 
It is evaluated on a video crowd-monitoring dataset \cite{chan2008privacy} in which each frame is first partitioned into homogeneous motion components via a mixture-of-dynamic-textures model; counts are then estimated per segment under a Poisson likelihood whose log-mean is regressed from hand-crafted low-level features. 
However, this pipeline presupposes motion segmentation and video cues that are unavailable in still-image crowd-counting benchmarks.
Also, the supervision is based on each coarse segment (counts per motion component rather than fine-grained blockwise labels), leading to underutilized location information and reduced model performance. 
In contrast, our method trains end-to-end on images with fine-grained, blockwise supervision and optimizes a zero-inflated Poisson objective \minorrevreplace{that separates structural zeros from positive counts}{with separate zero-inflation and Poisson components}, directly addressing label sparsity while remaining effective in dense scenes.

\section{Method}

Our proposed \textbf{ZIP} framework (illustrated in Fig.~\ref{fig:model}) is based on \textbf{Z}ero-\textbf{I}nflated \textbf{P}oisson regression to address the extreme sparsity of ground-truth density maps. 
Our ZIP framework bypasses Gaussian smoothing and models per-block counts $y \in \mathbb{N}$ via a zero‑inflated Poisson:
\begin{equation}
P(y = k \mid \pi, \lambda) = \begin{cases}
    \pi + (1 - \pi) e^{-\lambda}, \quad & k = 0 \\
    (1 - \pi) \cfrac{e^{-\lambda} \lambda^k}{k!}, \quad & k > 0
\end{cases}
\label{eqn:zip_pmf}
\end{equation}
with $\pi$, $\lambda$ predicted per block.
Building on Eq.~(\ref{eqn:zip_pmf}), we now specify how $\pi$ and $\lambda$ are parameterized and predicted per block, and how these outputs are coupled with EBC to yield a stable learning objective.

\begin{revblock}
We parameterize the Poisson rate $\lambda$ using an EBC-style bin-expectation head~\cite{ma2024clip}. 
This choice is motivated by the long-tailed and imbalanced nature of blockwise crowd counts: most blocks are empty, while a small fraction of positive blocks contain much larger counts. 
Predicting a distribution over integer-valued count bins provides a bounded and smoother parameterization of $\lambda$ than unconstrained direct regression, reducing ambiguity near count boundaries and improving optimization stability. The ZIP formulation itself is independent of this particular rate parameterization. 
In our final model, the bin-expectation head provides a stable estimate of $\lambda$, while the zero-inflation branch estimates \minorrevreplace{the structural-zero probability}{the zero-inflation probability} $\pi$.
Together they define the ZIP expectation $(1-\pi)\lambda$, which separates spatial suppression from count magnitude and supports the zero-map analyses in Sec.~\ref{sec:model_interpretation}.
\end{revblock}

Given an input image $\boldsymbol{X} \in \mathbb{R}^{3 \times H \times W}$, we first compute the feature map $\boldsymbol{f}$ using a backbone $\boldsymbol{F}$:
\begin{equation}
    \boldsymbol{f} = \boldsymbol{F} (\boldsymbol{X}) \in \mathbb{R}^{C \times h \times w}
    \label{eqn:extract_feat}
\end{equation}
where $h = H // r$, $w = W // r$ denote the spatial size of the output, and $r$ represents the block size.
Then $\boldsymbol{f}$ is passed through a $\boldsymbol{\lambda}$-head $\boldsymbol{H}_{\boldsymbol{\lambda}}$ (implemented as a $1\times 1$ convolution), followed by a softmax activation to generate a probability distribution over the $(n{-}1)$ \emph{positive} count bins:
\begin{equation}
    \boldsymbol{P}^*_{\boldsymbol{\lambda}} = \mathrm{Softmax} \left( \boldsymbol{H}_{\boldsymbol{\lambda}} (\boldsymbol{f}) \right) \in \mathbb{R}^{(n-1) \times  h \times w}.
    \label{eqn:prob_lambda_map}
\end{equation}
Note that we exclude the $\mathcal{B}_1 = \{0\}$ bin here, since the Poisson component assumes a strictly positive rate\footnote{Excluding $\{0\}$ from the positive-count bins does not mean that sampling zeros are not modeled. In fact, sampling zeros are modeled by the term $(1 - \pi) \exp (- \lambda)$ in Eq.~\eqref{eqn:zip_pmf}.} $\lambda > 0$.
We then compute the estimated Poisson rate map $\boldsymbol{\lambda} \in \mathbb{R}_+^{1 \times h \times w}$ as the expected value of the bin centers:
\begin{equation}
    \boldsymbol{\lambda}_{i,j} = \sum_{k=2}^{n} {\boldsymbol{P}^*_{\boldsymbol{\lambda}}}_{k,i,j} \cdot {b}_k,
    \label{eqn:lambda_map}
\end{equation}
where ${b}_k$ with $k = 2, \cdots, n$ denotes the center of each positive bin.

The learning of $\boldsymbol{P}^*_{\boldsymbol{\lambda}}$ is supervised by a cross-entropy loss over the positive regions of the ground-truth density map.
Specifically, let $\boldsymbol{Y}$ be the ground-truth blockwise density map without Gaussian smoothing, where $\boldsymbol{Y}_{i,j}$ denote the discrete blockwise count for block $(i,j)$. We define the set of positive-valued blocks as $\boldsymbol{Y}_+ \coloneqq \boldsymbol{Y}[\boldsymbol{Y} > 0]$, which arise from the Poisson component and are thus only related to $\boldsymbol{\lambda}$.
Their corresponding probabilistic predictions $\boldsymbol{P}^*_{\boldsymbol{\lambda}+}$ can be obtained via $\boldsymbol{P}^*_{\boldsymbol{\lambda}+} = \boldsymbol{P}^*_{\boldsymbol{\lambda}}[:, \boldsymbol{Y} > 0]$.
The cross-entropy loss is calculated via:
\begin{equation}
    \mathcal{L}_{\mathrm{CE}} = \mathrm{CrossEntropy}(\boldsymbol{P}^*_{\boldsymbol{\lambda}+}, \boldsymbol{P}_+),
    \label{eqn:ce_loss}
\end{equation}
where $\boldsymbol{P}_+$ is the ground-truth probability map one-hot encoded from $\boldsymbol{Y}_+$.

In parallel, we introduce a $\boldsymbol{\pi}$-head, $\boldsymbol{H}_{\boldsymbol{\pi}}$, also implemented as a $1 \times 1$ convolution, followed by a sigmoid activation to produce 
\minorrevreplace{
the structural zero probability map
}{
the zero-inflation probability map
}:
\begin{equation}
    \boldsymbol{\pi} = \sigma \left( \boldsymbol{H}_{\boldsymbol{\pi}} (\boldsymbol{f}) \right) \in \mathbb{R}^{1 \times  h \times w}.
    \label{eqn:pi_map}
\end{equation}

The negative log-likelihood (NLL) loss of the zero-inflated Poisson distribution is given by:
\begin{equation}
    \mathcal{L}_\mathrm{NLL} = - \frac{1}{hw} \sum_{i=1}^{h} \sum_{j=1}^{w} \log P(\boldsymbol{Y}_{i,j} \mid \boldsymbol{\pi}_{i,j}, \boldsymbol{\lambda}_{i,j} ),
    \label{eqn:nll_loss}
\end{equation}
where $P(\boldsymbol{Y}_{i,j} \mid \boldsymbol{\pi}_{i,j}, \boldsymbol{\lambda}_{i,j} )$ is the p.m.f. of the ZIP distribution given by Eq.~\eqref{eqn:zip_pmf}, and $\boldsymbol{\lambda}$ and $\boldsymbol{\pi}$ are given by \eqref{eqn:lambda_map} and \eqref{eqn:pi_map}, respectively.

It is important to note that we do not supervise the $\boldsymbol{\pi}$ and $\boldsymbol{\lambda}$ branches separately.
Since structural and sampling zeros are not explicitly annotated in the ground-truth, we treat them as latent factors and optimize both branches jointly via the ZIP negative log-likelihood loss in Eq.~\eqref{eqn:nll_loss}.
\minorrevreplace{
This formulation allows the model to implicitly learn to disentangle the two types of zeros based on the statistical patterns in the data, without requiring additional annotations.
}{
For a zero-count block $(i,j)$, the ZIP likelihood depends on
$\boldsymbol{\pi}_{i,j}$ and $\boldsymbol{\lambda}_{i,j}$ only through the zero probability
$(\boldsymbol{\pi}_{i,j}+(1-\boldsymbol{\pi}_{i,j})\exp(-\boldsymbol{\lambda}_{i,j}))$;
hence, the two contributions are not separately identifiable from
that observation alone.
Across the training data, positive-count likelihood terms penalize
large $\boldsymbol{\pi}_{i,j}$, $\mathcal{L}_{\mathrm{CE}}$ constrains
$\boldsymbol{\lambda}_{i,j}$ on positive blocks, and
$\mathcal{L}_{\mathrm{count}}$ constrains the summed expectation
$\sum_{i,j}(1-\boldsymbol{\pi}_{i,j})\boldsymbol{\lambda}_{i,j}$.
Together, these terms guide the joint estimation of the
$\boldsymbol{\pi}$ and $\boldsymbol{\lambda}$ maps, whose learned
behavior is further examined in Sec.~\ref{sec:model_interpretation}.
}

We use the expectation map $\boldsymbol{Y}^* = \mathbb{E}[\boldsymbol{Y} | \boldsymbol{\pi}, \boldsymbol{\lambda}]$ as the predicted density map, given by
\begin{equation}
    \boldsymbol{Y}^*
     = (1 - \boldsymbol{\pi}) \otimes \boldsymbol{\lambda} \in \mathbb{R}^{1 \times h \times w},
    \label{eqn:zip_density_map}
\end{equation}
where $\otimes$ represents element-wise multiplication.

Following DMCount \cite{wang2020distribution}, we incorporate a count loss that penalizes discrepancies between the predicted and ground-truth total counts to ensure accurate crowd estimates at the image level.
The predicted total count $c^*$ is obtained by summing over all elements of the predicted density map:
\begin{equation}
    c^* = \sum_{i=1}^h \sum_{j=1}^w \boldsymbol{Y}^*_{i,j}.
    \label{eqn:pred_count}
\end{equation}
We define the count loss as the MAE between the predicted and ground-truth total counts:
\begin{equation}
    \mathcal{L}_{\mathrm{count}} = \left| c^* - c \right| = \left| \sum_{i=1}^{h} \sum_{j=1}^{w} \left( \boldsymbol{Y}^*_{i,j} - \boldsymbol{Y}_{i,j} \right) \right|
    \label{eqn:count_loss}
\end{equation}

Finally, the model is trained using a weighted sum of three loss terms:
\begin{equation}
    \mathcal{L}_{\mathrm{total}} = \omega \mathcal{L}_{\mathrm{CE}} + \mathcal{L}_{\mathrm{NLL}} + \mathcal{L}_{\mathrm{count}},
    \label{eqn:total_loss}
\end{equation}
where $\omega$ is a scalar weighting coefficient, and $\mathcal{L}_{\mathrm{CE}}$, $\mathcal{L}_{\mathrm{NLL}}$, and $\mathcal{L}_{\mathrm{count}}$ are defined in Eq.~\eqref{eqn:ce_loss}, \eqref{eqn:nll_loss}, and \eqref{eqn:count_loss}, respectively.

\section{Generalization Analysis of ZIP}
\label{sec:risk_bound}

We analyze the generalization behavior of the proposed ZIP negative log-likelihood under bounded blockwise counts and compact parameter assumptions.
The goal is to establish a Rademacher generalization bound for ZIP NLL and clarify its relation to standard Poisson NLL and the OT loss used in DMCount~\cite{wang2020distribution}.
In particular, ZIP NLL and Poisson NLL have the same order of uniform generalization gap, while ZIP provides a richer likelihood family for zero-heavy blockwise count distributions.

Let $s \coloneqq hw$ denote the number of spatial blocks in an image.
For a single block with count $k$ and parameters $\theta \coloneqq (\pi,\lambda)$ in a compact domain
$\Theta \coloneqq [\pi_{\min},\pi_{\max}] \times [\lambda_{\min},\lambda_{\max}] \subset (0,1)\times[1,\Lambda]$, the ZIP negative log-likelihood is
\begin{equation}
    \ell_{\theta}^{\mathrm{ZIP}}(k)
    =
    \begin{cases}
        -\log[\pi+(1-\pi)e^{-\lambda}], & k=0, \\
        -\log(1-\pi)+\lambda-k\log\lambda+\log k!, & k\ge 1.
    \end{cases}
    \label{eqn:zip_block_nll}
\end{equation}
For the vector of blockwise counts $\boldsymbol{z}=(z_1,\ldots,z_s)$ and blockwise parameters $\boldsymbol{\theta}=(\theta_1,\ldots,\theta_s)$, define the full-image ZIP NLL as
\begin{equation}
    \mathcal{L}_{\mathrm{ZIP}}(\boldsymbol{z};\boldsymbol{\theta})
    =
    \frac{1}{s}\sum_{i=1}^{s}\ell_{\theta_i}^{\mathrm{ZIP}}(z_i).
    \label{eqn:zip_image_nll}
\end{equation}
The compactness assumption is standard in Rademacher analyses: in practice, $\pi$ can be kept away from the endpoints by bounded logits or clipping, and $\lambda$ is bounded because it is computed from finite bin centers in Eq.~\eqref{eqn:lambda_map}.

\begin{lemma}
    Under bounded counts $z_i\le k_{\max}$ and compact $\Theta$, the per-block ZIP NLL $\ell_{\theta}^{\mathrm{ZIP}}(k)$ is $l$-Lipschitz continuous with respect to the count variable $k$.
    \label{lem:1}
\end{lemma}

\begin{lemma}
    The full-image ZIP NLL $\mathcal{L}_{\mathrm{ZIP}}(\boldsymbol{z};\boldsymbol{\theta})$ is $L_{\mathrm{ZIP}}$-Lipschitz continuous with respect to the joint $\ell_1$ norm over $(\boldsymbol{z},\boldsymbol{\theta})$.
    \label{lem:2}
\end{lemma}

The proofs of Lemma~\ref{lem:1} and Lemma~\ref{lem:2} are provided in the Supplementary Material.
Following the Rademacher-complexity argument used in~\cite{wang2020distribution}, these Lipschitz properties yield the following bound.

\begin{theorem}
\label{thm:main}
Let $\mathcal{H}$ denote the base hypothesis class for predicting blockwise ZIP parameters, and define the full blockwise hypothesis class as $\mathcal{F}_{\mathrm{ZIP}}\coloneqq \mathcal{H}^{\times hw}$.
Let $f_{\mathcal S}^{\mathrm{ZIP}}$ be the empirical risk minimizer of $\mathcal{L}_{\mathrm{ZIP}}$ over $K$ \textit{i.i.d.} training samples, and let $f_{\mathcal D}^{\mathrm{ZIP}}$ be the corresponding population minimizer.
Then, for any $\delta\in(0,1)$, with probability at least $1-\delta$,
\begin{align}
    &\mathcal{R}_{\mathcal D}
    (f_{\mathcal S}^{\mathrm{ZIP}},\mathcal{L}_{\mathrm{ZIP}})
    -
    \mathcal{R}_{\mathcal D}
    (f_{\mathcal D}^{\mathrm{ZIP}},\mathcal{L}_{\mathrm{ZIP}})
    \notag\\
    &\quad \le
    2hw L_{\mathrm{ZIP}}\mathfrak{R}_{\mathcal S}(\mathcal H)
    +
    \mathcal{O}(\sqrt{1/K}),
    \label{eqn:zip_bound}
\end{align}
where $\mathfrak{R}_{\mathcal S}(\mathcal H)$ is the empirical Rademacher complexity of the base hypothesis class.
\end{theorem}

\begin{revblock}
\paragraph{Relation to Poisson NLL.}
For comparison, the standard Poisson NLL for a blockwise count $k$ is
\begin{equation}
    \ell_{\lambda}^{\mathrm{P}}(k)
    =
    \lambda-k\log\lambda+\log k!,
    \quad
    \lambda\in[\lambda_{\min},\lambda_{\max}].
    \label{eqn:poisson_block_nll}
\end{equation}
Under the same bounded-count and compact-parameter assumptions, the same argument gives
\begin{align}
    &\mathcal{R}_{\mathcal D}
    (f_{\mathcal S}^{\mathrm{P}},\mathcal{L}_{\mathrm{P}})
    -
    \mathcal{R}_{\mathcal D}
    (f_{\mathcal D}^{\mathrm{P}},\mathcal{L}_{\mathrm{P}})
    \notag\\
    &\quad \le
    2hw L_{\mathrm{P}}\mathfrak{R}_{\mathcal S}(\mathcal H)
    +
    \mathcal{O}(\sqrt{1/K}),
    \label{eqn:poisson_bound}
\end{align}
where $L_{\mathrm{P}}$ is the Lipschitz constant of the full-image Poisson NLL.
Thus, ZIP NLL and Poisson NLL have the same order of uniform generalization gap.
The advantage of ZIP is not a smaller worst-case estimation term, but a richer likelihood family: if the closure of the ZIP parameter space includes $\pi=0$, ZIP recovers the standard Poisson likelihood, while $\pi>0$ models excess zeros that cannot be represented by a standard Poisson distribution.
Consequently, the optimal ZIP population risk is no worse than the optimal Poisson population risk in the enlarged or limiting model class, and can be strictly lower when the conditional blockwise count distribution is genuinely zero-inflated.
The derivation of Eq.~\eqref{eqn:poisson_bound} is provided in the Supplementary Material.
\end{revblock}

\paragraph{Comparison with OT.}
For the OT loss in DMCount~\cite{wang2020distribution}, the generalization gap is bounded by
\begin{align}
    &\mathcal{R}_{\mathcal D}
    (f_{\mathcal S}^{\mathrm{OT}},\mathcal{L}_{\mathrm{OT}})
    -
    \mathcal{R}_{\mathcal D}
    (f_{\mathcal D}^{\mathrm{OT}},\mathcal{L}_{\mathrm{OT}})
    \notag\\
    &\quad \le
    4(hw)^2 C_{\infty}\mathfrak{R}_{\mathcal S}(\mathcal H)
    +
    \mathcal{O}(\sqrt{1/K}),
    \label{eqn:ot_bound}
\end{align}
where $C_{\infty}$ is the maximum cost in the OT cost matrix.
\minorrevreplace{
Therefore, ZIP NLL retains the same order of generalization gap as Poisson NLL while scaling more favorably than OT-based supervision with respect to the number of spatial blocks. This theoretical relation is consistent with the controlled empirical comparison among EBC, Poisson NLL, and ZIP NLL in Table~\ref{tab:ablation_full}.
}{
Therefore, under the stated assumptions, the ZIP NLL generalization bound has the same order as the Poisson NLL bound and a lower-order dependence on the number of spatial blocks than the OT-based bound.
}
The same argument can be extended to the full training objective in Eq.~\eqref{eqn:total_loss}; details are provided in the Supplementary Material.

%
\begin{table}[t]
\centering
\caption{ZIP variants with corresponding backbones, parameter sizes, and computational complexity measured at $1920 \times 1080$.}
\label{tab:ebc_zip_variants}
\begin{tabular}{l|l|c|c}
\toprule
\textbf{Suffix}  & \textbf{Backbone}                  & \textbf{Size (M)} & \textbf{FLOPs (G)} \\
\midrule
-P (Pico)  & MobileNetV4-Small-$0.5\times$ & 0.81                   & 6.46               \\
-N (Nano)  & MobileNetV4-Small             & 3.36                   & 24.73              \\
-T (Tiny)  & MobileNetV4-Medium            & 10.64                  & 61.62              \\
-S (Small) & MobileCLIP-S1                 & 33.72                  & 242.68             \\
-B (Base)  & CLIP-ConvNeXt-Base                 & 105.60                 & 800.99 \\         
\bottomrule
\end{tabular}%
\end{table}

\section{Experiments}

\noindent\textbf{Variants.} To evaluate the scalability and adaptability of our proposed framework, we construct five variants of ZIP with different backbones, including MobileNetV4 \cite{qin2024mobilenetv4}, MobileCLIP \cite{vasu2024mobileclip} and OpenCLIP's ConvNeXt-Base \cite{liu2022convnet, cherti2023reproducible}. These variants are suitable for different application scenarios, ranging from mobile-friendly to high-performance settings. The specifications are summarized in Table~\ref{tab:ebc_zip_variants}. All variants share the same head design and differ only in backbone complexity. This modularity allows ZIP to scale across computational budgets.

\begin{table*}[t]
\centering
\minorrev{
\caption{
Controlled ablation using the same VGG19-based backbone and training protocol. 
We vary blockwise supervision, density parameterization, cross-entropy supervision, zero-inflation, and image-level count loss. 
Direct-regression variants predict non-negative counts/rates via Softplus and do not use $\mathcal{L}_{\mathrm{CE}}$; bin-expectation variants estimate counts from expected bin centers. 
All results are reported as mean $\pm$ standard deviation over three independent runs initialized with seeds 42, 3407, and 2026.
}
\label{tab:ablation_full}
\resizebox{\textwidth}{!}{
\begin{tabular}{c|l|l|c|c|c|c|cc|cc|cc}
\toprule
\multirow{2}{*}{\textbf{Exp.}} &
\multirow{2}{*}{\textbf{Blockwise Term}} &
\multirow{2}{*}{\textbf{Density Param.}} &
\multirow{2}{*}{\textbf{Binning}} &
\multirow{2}{*}{$\boldsymbol{\mathcal{L}_{\mathrm{CE}}}$} &
\multirow{2}{*}{\textbf{Zero Inflation}} &
\multirow{2}{*}{$\boldsymbol{\mathcal{L}_{\mathrm{count}}}$} &
\multicolumn{2}{c|}{\textbf{ShanghaiTech A}} &
\multicolumn{2}{c|}{\textbf{ShanghaiTech B}} &
\multicolumn{2}{c}{\textbf{UCF-QNRF}} \\
 & & & & & & & \textbf{MAE} & \textbf{RMSE} & \textbf{MAE} & \textbf{RMSE} & \textbf{MAE} & \textbf{RMSE} \\
\midrule
\multicolumn{13}{l}{\textit{A. Direct regression baselines}} \\
A1 & MAE         & Direct Regression & -- & N/A & -- & \checkmark & $63.5 \pm 0.7$ & $101.5 \pm 1.1$ & $7.0 \pm 0.1$ & $11.2 \pm 0.2$ & $82.4 \pm 0.5$ & $143.1 \pm 3.8$ \\
A2 & MSE         & Direct Regression & -- & N/A & -- & \checkmark & $65.3 \pm 0.9$ & $103.0 \pm 2.5$ & $7.5 \pm 0.0$ & $12.3 \pm 0.1$ & $84.1 \pm 1.7$ & $147.9 \pm 3.4$ \\
A3 & DMCount     & Direct Regression & -- & N/A & -- & \checkmark & $62.3 \pm 0.4$ & $99.7 \pm 2.5$ & $7.0 \pm 0.1$ & $11.2 \pm 0.2$ & $80.5 \pm 0.2$ & $138.5 \pm 2.7$ \\
A4 & Poisson NLL & Direct Regression & -- & N/A & -- & \checkmark & $62.0 \pm 0.3$ & $101.4 \pm 2.2$ & $6.7 \pm 0.1$ & $10.6 \pm 0.1$ & $77.0 \pm 1.3$ & $131.4 \pm 4.2$ \\
A5 & ZIP NLL     & Direct Regression & -- & N/A & \checkmark & \checkmark & $\underline{59.8} \pm 0.6$ & $\underline{96.3} \pm 2.0$ & $6.3 \pm 0.1$ & $\underline{10.1} \pm 0.2$ & $\underline{74.2} \pm 0.8$ & $129.8 \pm 1.9$ \\
\midrule
\multicolumn{13}{l}{\textit{B. Bin-expectation framework comparison}} \\
B1 & MAE         & Bin Expectation & \checkmark & \checkmark & -- & \checkmark & $62.6 \pm 1.0$ & $99.9 \pm 3.2$ & $6.4 \pm 0.0$ & $10.7 \pm 0.3$ & $80.4 \pm 0.7$ & $139.0 \pm 2.2$ \\
B2 & MSE         & Bin Expectation & \checkmark & \checkmark & -- & \checkmark & $62.7 \pm 0.9$ & $101.5 \pm 1.1$ & $6.9 \pm 0.1$ & $11.5 \pm 0.2$ & $82.9 \pm 1.1$ & $144.7 \pm 3.7$ \\
B3 & DMCount     & Bin Expectation & \checkmark & \checkmark & -- & \checkmark & $62.1 \pm 0.8$ & $102.3 \pm 2.1$ & $6.6 \pm 0.1$ & $11.3 \pm 0.2$ & $76.4 \pm 1.0$ & $130.8 \pm 4.2$ \\
B4 & Poisson NLL & Bin Expectation & \checkmark & \checkmark & -- & \checkmark & $60.8 \pm 1.1$ & $98.8 \pm 2.6$ & $\underline{6.2} \pm 0.0$ & $10.3 \pm 0.1$ & $76.5 \pm 0.7$ & $\underline{129.4} \pm 1.7$ \\
B5 & ZIP NLL     & Bin Expectation & \checkmark & \checkmark & \checkmark & \checkmark & $\mathbf{58.9} \pm 0.8$ & $\mathbf{94.3} \pm 1.3$ & $\mathbf{6.0} \pm 0.1$ & $\mathbf{9.9} \pm 0.1$ & $\mathbf{72.6} \pm 0.9$ & $\mathbf{124.8} \pm 2.6$ \\
\midrule
\multicolumn{13}{l}{\textit{C. Component ablation of ZIP (B5)}} \\
C1 & ZIP NLL     & Bin Expectation & \checkmark & -- & \checkmark & \checkmark & $64.9 \pm 0.9$ & $103.5 \pm 2.0$ & $6.9 \pm 0.1$ & $11.0 \pm 0.2$ & $84.9 \pm 1.8$ & $145.7 \pm 0.9$ \\
C2 & ZIP NLL     & Bin Expectation & \checkmark & \checkmark & \checkmark & -- & $62.1 \pm 0.8$ & $100.7 \pm 1.5$ &  $6.4 \pm 0.1$ & $10.6 \pm 0.2$ & $79.1 \pm 1.0$ & $138.0 \pm 2.8$ \\
\bottomrule
\end{tabular}
}
}
\end{table*}
\begin{figure}[t]
    \centering
    \includegraphics[width=\linewidth]{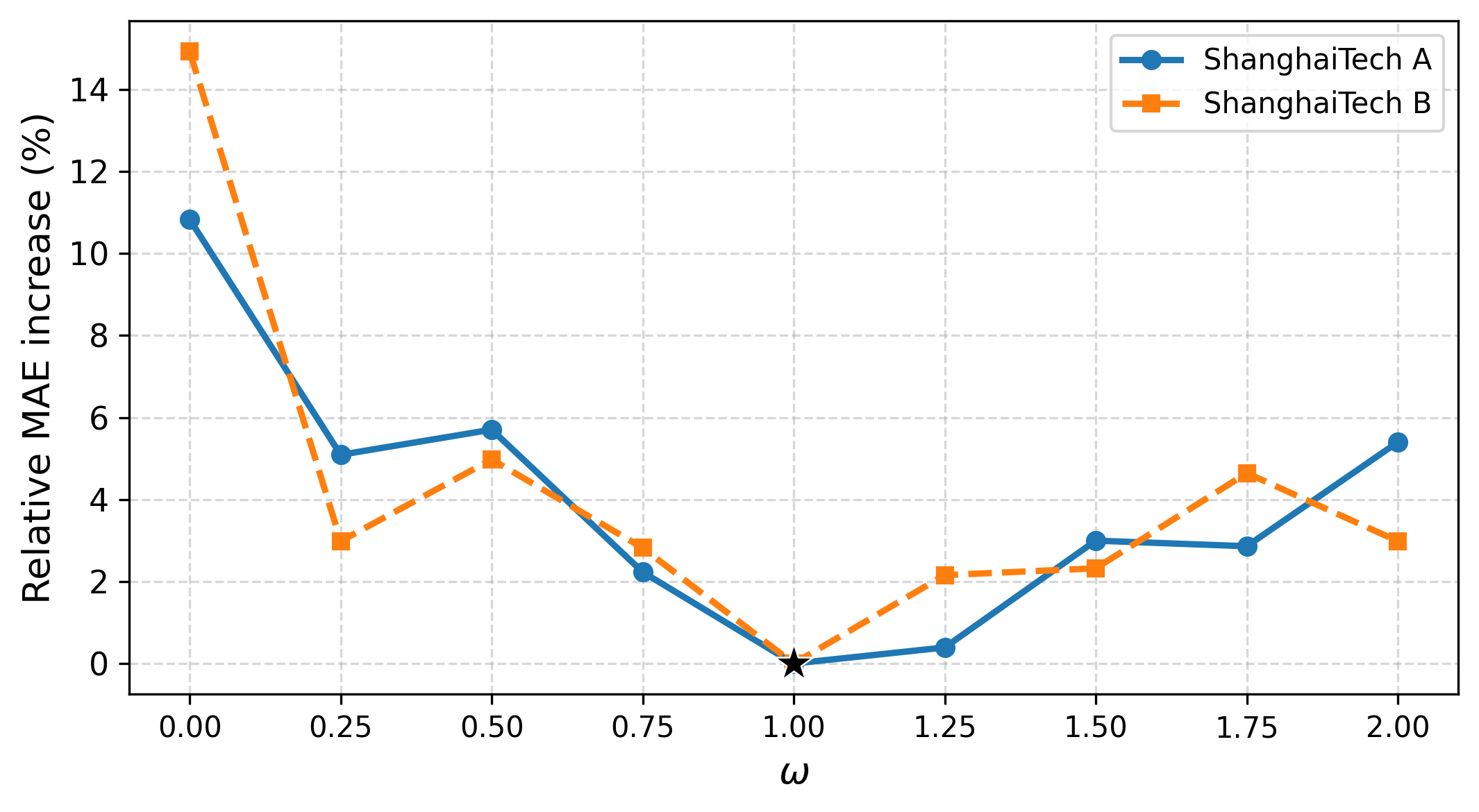}
    \caption{\rev{
    Sensitivity of MAE to the CE weight $\omega$ in Eq.~\eqref{eqn:total_loss} on ShanghaiTech A and B. 
    For each dataset, MAE is normalized as the relative increase over the best value observed across all $\omega$ values. 
    The black star marks the shared best setting at $\omega=1.00$, where the MAE is 58.74 on ShanghaiTech A and 6.03 on ShanghaiTech B.
    }}
    \label{fig:omega}
\end{figure}

\noindent\textbf{Setup.} Our training configuration largely follows EBC.
We first preprocess all datasets so that the minimum edge length is no less than 448. We limit the maximum edge length of UCF-QNRF to be 2048 and that of NWPU-Crowd to be 3072. 
We apply a combination of data augmentation techniques, including random resized cropping, horizontal flipping, and color jittering (see supplementary material for more details).
Block sizes are set to 16 for ShanghaiTech A \& B and NWPU-Crowd, and 32 for UCF-QNRF.

All models are optimized using the Adam optimizer \cite{kingma2014adam} with a weight decay of $1e-4$.
The learning rate is linearly warmed up from $1e-5$ to $1e-4$ over the first 25 epochs, followed by cosine annealing with parameters $T_0 = 5$ and $T_\mathrm{mult} = 2$.
All models were trained for 660 epochs using PyTorch 2.7.1 and Ubuntu 24.04 LTS.

\noindent\textbf{Evaluation metrics.} Following standard practice, we evaluate each model variant on all datasets using three commonly adopted metrics: MAE, RMSE, and Normalized Absolute Error (NAE). 
NAE reflects the relative error of predictions by normalizing with respect to the ground-truth count. 
Compared with MAE and RMSE, it reduces the bias towards highly populated scenes, thereby serving as a fairer metric across images with different crowd densities.

\begin{table*}[t]
\caption{
\rev{
System-level comparison with reported results of recent state-of-the-art crowd-counting methods on ShanghaiTech A \& B, UCF-QNRF, and the NWPU-Crowd validation split. ZIP-B achieves the best reported MAE on all four datasets and the best reported RMSE on ShanghaiTech A, UCF-QNRF, and NWPU-Crowd Val. Inference FLOPs are measured at $1920 \times 1080$.
}
}
\label{tab:comparison_sota}
\centering
\begin{tabular}{l|cc|cccccccc}
\toprule
\multirow{2}{*}{\textbf{Method}} 
& \multirow{2}{*}{\rev{\textbf{Size (M)}}} 
& \multirow{2}{*}{\rev{\textbf{FLOPs (G)}}} 
& \multicolumn{2}{c}{\textbf{ShanghaiTech A}} 
& \multicolumn{2}{c}{\textbf{ShanghaiTech B}} 
& \multicolumn{2}{c}{\textbf{UCF-QNRF}} 
& \multicolumn{2}{c}{\textbf{NWPU (Val)}} \\
& & 
& \textbf{MAE} & \textbf{RMSE} 
& \textbf{MAE} & \textbf{RMSE} 
& \textbf{MAE} & \textbf{RMSE} 
& \textbf{MAE} & \textbf{RMSE} \\ 
\midrule
ChfL \cite{shu2022crowd}             
& \rev{21.5} & \rev{853.7} & 57.5 & 94.3 & 6.9 & 11.0 & 80.3 & 137.6 & 76.8 & 343.0 \\
MAN \cite{lin2022boosting}           
& \rev{40.4} & \rev{1,152.1} & 56.8 & 90.3 & -- & -- & 77.3 & 131.5 & 76.5 & 323.0 \\
CLTR \cite{liang2022end}             
& \rev{43.5} & \rev{218.0} & 56.9 & 95.2 & 6.5 & 10.6 & 85.8 & 141.3 & 61.9 & 246.3 \\
CrowdHat \cite{wu2023boosting}       
& \rev{38.0} & \rev{3,188.7} & 51.2 & 81.9 & 5.7 & 9.4 & 75.1 & \underline{126.7} & -- & -- \\
STEERER \cite{han2023steerer}        
& \rev{65.5} & \rev{745.6} & 54.5 & 86.9 & 5.8 & \textbf{8.5} & 74.3 & 128.3 & 54.3 & 238.3 \\
PET \cite{liu2023point}              
& \rev{57.9} & \rev{960.8} & 49.3 & 78.8 & 6.2 & 9.7 & 79.5 & 144.3 & 58.5 & 238.0 \\
APGCC \cite{chen2024improving}       
& \rev{19.6} & \rev{751.7} & \underline{48.8} & \underline{76.7} & \underline{5.6} & 8.7 & 80.1 & 133.6 & -- & -- \\
Gramformer \cite{lin2024gramformer}  
& \rev{29.0} & \rev{942.8} & 54.7 & 87.1 & -- & -- & 76.7 & 129.5 & -- & -- \\
mPrompt \cite{guo2024regressor}      
& \rev{48.3} & \rev{3,371.6} & 52.5 & 88.9 & 5.8 & 9.6 & \underline{72.2} & 133.1 & 50.2 & 219.0 \\
CLIP-EBC \cite{ma2024clip}           
& \rev{97.1} & \rev{978.0} & 52.5 & 85.9 & 6.6 & 10.5 & 80.3 & 139.3 & \underline{36.6} & \underline{81.7} \\ 
\midrule
\textbf{ZIP-B (ours)}                
& \rev{105.6} & \rev{800.9}
& \textbf{47.8} & \textbf{75.0} 
& \textbf{5.5} & \underline{8.6} 
& \textbf{69.4} & \textbf{121.8} 
& \textbf{28.2} & \textbf{64.8} \\ 
\bottomrule
\end{tabular}
\end{table*}

\begin{table*}[t]
\caption{Comparison of our method ZIP-B with the latest approaches of comparable scales on the test split of NWPU-Crowd. ZIP-B achieves both the lowest MAE and NAE, reducing NAE by 21.8\% compared to the previous best (STEERER).}
\label{tab:comparison_sota_nwpu}
\centering
\begin{tabular}{l|ccc|C{0.9cm}C{0.9cm}C{0.9cm}|cccc}
\toprule
\multirow{2}{*}{\textbf{Method}}    & \multirow{2}{*}{\textbf{MAE}}  & \multirow{2}{*}{\textbf{RMSE}}  & \multirow{2}{*}{\textbf{NAE (\%)}} & \multicolumn{3}{c|}{\textbf{Luminance-Level NAE (\%)}}            & \multicolumn{4}{c}{\textbf{Scene-Level NAE (\%)}}                                  \\ 
          & & & & \textbf{L1}             & \textbf{L2}             & \textbf{L3}            & \textbf{S1}             & \textbf{S2}            & \textbf{S3}             & \textbf{S4}             \\ \midrule
ChfL \cite{shu2022crowd} & 76.8          & 343.0          & 17.1  & 21.1 & 17.8 & 13.6 & 23.2 & 13.8 & 14.6 & 25.9 \\
MAN \cite{lin2022boosting} & 76.5          & 323.0          & 17.0  & 21.5 & 17.0 & 15.8 & 20.3 & 15.2 & 15.5 & 28.10 \\
CLTR \cite{liang2022end} & 74.4          & 333.8          & 16.5 & 20.4 & 17.1 & 13.1 & 22.5 & 12.9 & 14.3 & 32.6 \\
CrowdHat \cite{wu2023boosting} & 68.7          & 296.9          & 18.1 & 19.1          & 19.6          & 13.0         & 19.3          & 18.6         & 14.6          & \underline{21.6}    \\
STEERER \cite{han2023steerer}  & 63.7          & 309.8          & \underline{13.3} & \textbf{14.3} & \underline{14.2}    & \underline{9.9}    & \underline{16.6}    & \underline{11.2}   & 12.2          & 23.6          \\
PET \cite{liu2023point} & 74.4          & 328.5          & - & - & - & - & - & - & - & - \\
APGCC \cite{chen2024improving} & 71.7          & \underline{284.4}    & 21.5 & 31.5          & 22.0          & 17.2         & 36.1          & 14.7         & 14.1          & 29.3          \\
Gramformer \cite{lin2024gramformer} & 72.5          & 316.4          & 16.0 & - & - & - & - & - & - & - \\
mPrompt \cite{guo2024regressor} & 62.1          & 293.5          & -  & - & - & - & - & - & - & - \\
CLIP-EBC \cite{ma2024clip}  & \underline{61.3}    & \textbf{278.4} & 14.8 & 16.4          & 15.5          & 11.8         & 20.6          & 11.8         & \underline{12.1}    & 22.1          \\
\textbf{ZIP-B (ours)} & \textbf{60.1} & 299.0          & \textbf{10.4}    & \underline{14.4}    & \textbf{11.2} & \textbf{8.2} & \textbf{11.1} & \textbf{9.9} & \textbf{11.6} & \textbf{21.5} \\ \bottomrule
\end{tabular}
\end{table*}

\subsection{Ablation Study}

\rev{
We first analyze the sensitivity to the weighting parameter $\omega$ in Eq.~\eqref{eqn:total_loss}, which controls the contribution of the cross-entropy term relative to the likelihood and image-level count losses. 
To this end, we evaluate the VGG19-based encoder-decoder model under different values of $\omega$ on ShanghaiTech A and ShanghaiTech B. 
Fig.~\ref{fig:omega} reports the relative MAE increase over the best value observed for each dataset. 
Both datasets achieve their lowest MAE at $\omega=1.00$, with MAE values of 58.7 on ShanghaiTech A and 6.0 on ShanghaiTech B. 
Therefore, we use $\omega=1.00$ as the default setting for all subsequent experiments without dataset-specific tuning.
}

\rev{
To validate the contribution of each modeling choice, we conduct a controlled ablation on ShanghaiTech A, ShanghaiTech B, and UCF-QNRF, as shown in Table~\ref{tab:ablation_full}. 
The ablation is organized into three groups. 
Group A removes the EBC-style bin-expectation parameterization and directly regresses non-negative blockwise density maps through a Softplus output. 
Since no categorical distribution over count bins is produced in these direct-regression variants, $\mathcal{L}_{\mathrm{CE}}$ is not applicable. 
Within this group, A3 corresponds to DMCount~\cite{wang2020distribution}, implemented under the same controlled training protocol as the other variants.
Group B uses the same bin-expectation parameterization and compares different blockwise density supervision terms under identical binning, CE supervision, image-level count loss, and VGG19-based backbone. 
In this group, B3 is the controlled EBC baseline under our implementation: it uses the same bin-expectation parameterization and positive-bin CE supervision as EBC, together with the DMCount density-map loss and the image-level count loss. 
Thus, the comparison B3 $\rightarrow$ B5 isolates the effect of replacing the EBC density objective with the proposed ZIP likelihood while keeping binning, CE supervision, count loss, and backbone fixed. 
B4 is the standard Poisson NLL baseline, and B5 is the full proposed ZIP model.
Group C further ablates the two auxiliary terms in Eq.~\eqref{eqn:total_loss} around B5 by removing $\mathcal{L}_{\mathrm{CE}}$ or $\mathcal{L}_{\mathrm{count}}$ separately.
}

\rev{
Several observations can be made from Table~\ref{tab:ablation_full}. 
\minorrevreplace{First, ZIP NLL consistently outperforms Poisson NLL under both parameterizations.}{First, ZIP NLL achieves lower mean MAE and RMSE than Poisson NLL under both parameterizations on all three datasets.}
\minorrevreplace{The direct-regression variants in Group A show that the proposed ZIP formulation is not tied to EBC-style binning: even without binning or CE supervision, A5 consistently improves over direct-regression Poisson NLL (A4) and DMCount (A3) on all three datasets, with maximum relative gains of 6.0\% and 10.0\%, respectively.}{The direct-regression variants in Group A show that the proposed ZIP formulation is not tied to EBC-style binning: even without binning or CE supervision, A5 achieves lower mean MAE and RMSE than direct-regression Poisson NLL (A4) and DMCount (A3) on all three datasets, with maximum relative MAE reductions of 6.0\% and 10.0\%, respectively.}
\minorrevreplace{Under the final bin-expectation setting, B5 also improves over B4 on all datasets.}{Under the final bin-expectation setting, B5 likewise achieves lower mean MAE and RMSE than B4 on all three datasets.}
Since B4 and B5 use the same binning, CE supervision, count loss, and backbone, this comparison isolates the contribution of the zero-inflation branch.
}

\rev{
Second, B5 achieves lower errors than the EBC baseline B3 on all three datasets. 
\minorrevreplace{Specifically, B5 reduces MAE by 5.5\%, 9.1\%, and 5.1\% on ShanghaiTech A, ShanghaiTech B, and UCF-QNRF, respectively, and reduces RMSE by 7.6\%, 12.4\%, and 3.8\%.}{Specifically, B5 reduces mean MAE by 5.2\%, 9.1\%, and 5.0\% on ShanghaiTech A, ShanghaiTech B, and UCF-QNRF, respectively, and mean RMSE by 7.8\%, 12.4\%, and 4.6\%.}
This indicates that the proposed ZIP likelihood provides additional benefit over the EBC objective under the same bin-expectation framework.
}

\rev{
Finally, the component ablation in Group C shows that both auxiliary terms in Eq.~\eqref{eqn:total_loss} are useful. 
\minorrevreplace{Removing $\mathcal{L}_{\mathrm{CE}}$ from ZIP, \ie B5 $\rightarrow$ C1, causes consistent degradation on all datasets, suggesting that CE provides important optimization guidance for learning the positive-count bin distribution.}{Removing $\mathcal{L}_{\mathrm{CE}}$ from ZIP, \ie B5 $\rightarrow$ C1, increases the mean MAE and RMSE on all datasets, suggesting that CE provides important optimization guidance for learning the positive-count bin distribution.}
\minorrevreplace{Removing the image-level count loss, \ie B5 $\rightarrow$ C2, slightly degrades performance, indicating that $\mathcal{L}_{\mathrm{count}}$ contributes global count calibration.}{Removing the image-level count loss, \ie B5 $\rightarrow$ C2, also increases the mean errors, indicating that $\mathcal{L}_{\mathrm{count}}$ contributes global count calibration.}
}

\begin{table*}[t]
\caption{Comparison of our lightweight models with state-of-the-art lightweight approaches on ShanghaiTech A \& B, and UCF-QNRF. Models are grouped by parameter size: $<1$ M, $1-10$ M, $10 - 20$ M, $\ge 20$ M. FLOPs are measured on $1920 \times 1080$ resolution. The best result in each group is \ghl{highlighted}, the overall best result is shown in \textbf{bold}.}
\label{tab:comparison_sota_light}
\centering
\begin{tabular}{l|cc|cccccc}
\toprule
                                  &                                     &                                      & \multicolumn{2}{c}{\textbf{ShanghaiTech A}}                                  & \multicolumn{2}{c}{\textbf{ShanghaiTech B}}                             & \multicolumn{2}{c}{\textbf{UCF-QNRF}}                                      \\
\multirow{-2}{*}{\textbf{Method}} & \multirow{-2}{*}{\textbf{Size (M)}} & \multirow{-2}{*}{\textbf{FLOPs (G)}} & \textbf{MAE}                        & \textbf{RMSE}                          & \textbf{MAE}                       & \textbf{RMSE}                      & \textbf{MAE}                        & \textbf{RMSE}                        \\ \midrule
LSANet \cite{zhu2022real}                           & 0.2                                & 14.4                                & \ghl{66.1}                                & 110.2                                  & 8.6                                & 13.9                               & 112.3                               & 186.9                                \\
\textbf{ZIP-P (ours)}         & 0.8                                & 6.4                                 & 71.1                               & \ghl{109.6}          & \ghl{8.2}       & \ghl{12.6}      & \ghl{96.2}       & \ghl{161.8}       \\ \midrule \midrule
\textbf{ZIP-N (ours)}         & 3.3                                & 24.7                                & \ghl{58.8}       & \ghl{94.6}          & \ghl{7.7}       & \ghl{12.1}      & \ghl{86.4}       & \ghl{147.6}       \\
MobileCount \cite{wang2020mobilecount}                       & 3.4                                & 16.5                                & 89.4                                & 146.0                                  & 9.0                                & 15.4                               & 131.1                               & 222.6                                \\
LMSFFNet-XS \cite{yi2023lightweight}                      & 4.5                                & 39.3                                & 85.8                                & 139.9                                  & 9.2                                & 15.1                               & 112.8                               & 201.6                                \\
MobleCount$\times 1.25$ \cite{wang2020mobilecount}           & 5.5                                & 34.1                               & 82.9                                & 137.9                                  & 8.2                                & 13.2                               & 124.5                               & 207.6                                \\ \midrule \midrule
\textbf{ZIP-T (ours)}         & 10.6                               & 61.6                               & \ghl{56.3} & \ghl{90.5} & \ghl{6.6} & \ghl{9.9} & \ghl{76.0} & \ghl{129.4} \\
MobleCount$\times 2$ \cite{wang2020mobilecount}              & 13.4                               & 63.0                                & 81.4                                & 133.3                                  & 8.1                                & 12.7                               & 117.9                               & 207.5                                \\ \midrule \midrule
\textbf{ZIP-S (ours)}                  & 33.7                               & 242.7                               & \textbf{55.1}                      & \textbf{88.9}                            & \textbf{5.8}                      & \textbf{9.2}                      & \textbf{73.3}                      & \textbf{125.0}                      \\ \bottomrule
\end{tabular}
\end{table*}

\subsection{ZIP-B Compared with State-of-the-Art}

We first compare our base model, {ZIP-B}, with recent state-of-the-art crowd counting methods on four benchmark datasets: ShanghaiTech A \& B, UCF-QNRF, and the validation split of NWPU-Crowd. 
Table~\ref{tab:comparison_sota} summarizes the counting results, 
\rev{
model sizes, and computational complexity measured by inference FLOPs at 1920$\times$1080.
}

ZIP-B achieves the lowest MAE on all four datasets. 
In particular, on {ShanghaiTech A}, our model achieves the lowest MAE ({47.8}) and RMSE ({75.0}), outperforming the previous best result (48.8 MAE, 76.7 RMSE by APGCC).
On {ShanghaiTech B}, ZIP-B achieves the lowest MAE ({5.5}) and the second-best RMSE (\underline{8.6}), which is close to the best RMSE of STEERER (8.5).
For the challenging {UCF-QNRF} dataset, our method sets a new state-of-the-art with {69.4} MAE and {121.8} RMSE.
On {NWPU-Crowd (val)}, we obtain a substantial improvement over prior methods, achieving {28.2} MAE and {64.8} RMSE, significantly outperforming the closest competitor, CLIP-EBC, which reports 36.6 MAE and 81.7 RMSE.

We further evaluate our method on the NWPU-Crowd test set and stratify the results by both luminance levels and scene density levels. 
Luminance was measured using the average pixel intensity in the Y channel of the YUV color space, and divided into three groups: L1 $[0, 0.25]$, L2 $(0.25, 0.5]$, and L3 $(0.5, 0.75]$.
Scene-level difficulty was assessed based on the ground-truth number of people per image, grouped into four intervals: S1 $(0, 100]$, S2 $(100, 500]$, S3 $(500, 5000]$ and S4 ($> 5000$). 
Table~\ref{tab:comparison_sota_nwpu} presents the details.
ZIP-B achieves the lowest MAE ({60.1}) and lowest NAE ({10.4\%}).
Compared to the previous best-performing models, MAE is reduced by 1.2 (from 61.3 by CLIP-EBC), and NAE is improved by 21.8\% relative to STEERER (from 13.3\% to 10.4\%).
While the RMSE (299.0) is not the lowest among all methods, it remains competitive. 
Given that RMSE is highly sensitive to annotation noise (as illustrated in the supplementary material), MAE and NAE serve as more stable and representative metrics.
As for the luminance-level and scene-level comparison, ZIP-B outperforms prior methods across six out of seven subgroups, achieving the lowest NAE in L2 (11.2\%), L3 (8.2\%), and all four scene levels (S1--S4).
These results show that ZIP-B remains robust across varying lighting conditions and crowd densities. 
In particular, its gains under sparse scenes (S1) support the effectiveness of zero-inflated modeling.
%
%
%

\rev{
Overall, ZIP-B provides a strong system-level trade-off between accuracy and computational cost. 
Although it has a larger parameter count than the compared methods, its inference FLOPs are lower than several strong baselines, including MAN, CrowdHat, PET, mPrompt, and CLIP-EBC, while achieving the best MAE across all four benchmarks.
}

\begin{table*}[t]
\caption{Comparison of ZIP with the reported results of other frameworks on ShanghaiTech A \& B, UCF-QNRF and NWPU-Crowd. All frameworks use the same VGG19 encoder-decoder. ZIP achieves the best MAE/RMSE on all benchmarks and improves EBC by up to 14.3\%. Best and second-best results are highlighted in \textbf{bold} and \underline{underlined}, respectively.}
\label{tab:comparison_loss}
\centering
\begin{tabular}{l|cccccccc}
\toprule
\multirow{2}{*}{\textbf{Method}} & \multicolumn{2}{c}{\textbf{ShanghaiTech A}} & \multicolumn{2}{c}{\textbf{ShanghaiTech B}} & \multicolumn{2}{c}{\textbf{UCF-QNRF}} & \multicolumn{2}{c}{\textbf{NWPU-Crowd (Val)}} \\
                                 & \textbf{MAE}          & \textbf{RMSE}        & \textbf{MAE}         & \textbf{RMSE}         & \textbf{MAE}      & \textbf{RMSE}      & \textbf{MAE}          & \textbf{RMSE}         \\ \midrule
BL \cite{ma2019bayesian}                               & 62.8                  & 101.8                & 7.7                  & 12.7                  & 88.7              & 154.8              & 93.6                  & 470.3                 \\
DMCount \cite{wang2020distribution}                          & \underline{59.7}            & {95.7}        & 7.4                  & 11.8                  & 85.6              & 148.3              & 70.5                  & 357.6                 \\
GL \cite{wan2021generalized} & 61.3 & \underline{95.4} & 7.3 & 11.7 & 84.3 & 147.5 & - & - \\
EBC \cite{ma2024clip}                             & 62.3                  & 98.9                 & \underline{7.0}            & \underline{10.9}            & \underline{77.2}        & \underline{130.4}        & \underline{39.6}            & \underline{95.8}            \\ \midrule
\textbf{ZIP (ours)}          & \textbf{58.7}         & \textbf{93.6}           & \textbf{6.0}         & \textbf{9.9}         & \textbf{73.0}     & \textbf{126.1}     & \textbf{38.3}         & \textbf{90.3}         \\
Improvement over EBC             & 5.8\%                  & 5.4\%                 & 14.3\%                & 9.2\%                  & 5.4\%              & 3.3\%               & 3.3\%                  & 5.7\%                  \\ \bottomrule
\end{tabular}
\end{table*}

\begin{table*}[t]
\centering
\begin{revblock}
\caption{
\minorrevreplace{
Empirical validation of the resolution-dependent behavior on UCF-QNRF using the same VGG19-based encoder-decoder backbone and block size 32.
}{
Accuracy and training-time comparison across different training resolutions on UCF-QNRF using the same VGG19-based encoder-decoder backbone and block size 32.
}
The number of spatial blocks is $s=h\times w$.
Epoch time is measured for pure training over 12 repetitions; the highest and lowest measurements are discarded before computing the mean and standard deviation.
}
\label{tab:res_block_size}
\begin{tabular}{c|c|ccc|ccc}
\toprule
\multirow{2}{*}{\textbf{Training Res.}} &
\multirow{2}{*}{$\boldsymbol{s=h\times w}$} &
\multicolumn{3}{c|}{\textbf{MAE}} &
\multicolumn{3}{c}{\textbf{Epoch Time (s)}} \\
& & \textbf{DMCount (EBC)} & \textbf{ZIPNLL (ours)} & \textbf{Rel. Gain} & \textbf{DMCount (EBC)} & \textbf{ZIPNLL (ours)} & \textbf{Speedup} \\
\midrule
224 & $7\times7$   & 79.1 & 78.9 & 0.3\% & $29.1 \pm 0.1$ & $10.7 \pm 0.1$ & $2.7\times$ \\
448 & $14\times14$ & 78.1 & 75.7 & 3.1\% & $39.2 \pm 0.1$ & $16.3 \pm 0.1$ & $2.4\times$ \\
672 & $21\times21$ & 76.9 & \textbf{73.0} & \textbf{5.1\%} & $56.2 \pm 0.1$ & $32.8 \pm 0.1$ & $1.7\times$ \\
896 & $28\times28$ & 78.3 & 75.9 & 3.1\% & $80.7 \pm 0.2$ & $57.0 \pm 0.2$ & $1.4\times$ \\
\bottomrule
\end{tabular}%
\end{revblock}
\end{table*}

\subsection{Comparison with Lightweight Models}

To investigate the performance-efficiency trade-off, we compare our proposed lightweight models (from ZIP-P to ZIP-S
) with a range of representative compact methods.
%
We divide models into four groups based on parameter size: (1) ultra-lightweight models ($<$1M), (2) lightweight models (1–10M), (3) mid-size models (10–20M), and (4) larger lightweight models ($\ge$20M).
Table~\ref{tab:comparison_sota_light} presents the details.

\textbf{Ultra-lightweight ($<$1M).}
In this group, ZIP-P achieves the best results on five of six reported metrics: it improves RMSE on ShanghaiTech A and both MAE/RMSE on ShanghaiTech B and UCF-QNRF, while LSANet has lower MAE on ShanghaiTech A. This demonstrates our model's superior design in extremely constrained settings.

\textbf{Lightweight (1–10M).}
ZIP-N significantly outperforms other methods such as MobileCount and LMSFFNet-XS across all datasets. It achieves the lowest MAE in this group on ShanghaiTech A (58.8), ShanghaiTech B (7.7), and UCF-QNRF (86.4).

\textbf{Mid-size (10–20M).}
ZIP-T leads this group with the best results on all benchmarks. It achieves group-best scores of 56.3 MAE on ShanghaiTech A, 6.6 on ShanghaiTech B, and 76.0 on UCF-QNRF, significantly outperforming MobileCount$\times$2 with fewer parameters and FLOPs.

\textbf{Larger lightweight ($\ge$20M).}
ZIP-S sets a new benchmark in this group and across all lightweight models, achieving 55.1 MAE on ShanghaiTech A, 5.8 on ShanghaiTech B, and 73.3 on UCF-QNRF. These scores represent the best results among all compared lightweight models, and they also approach or surpass those of regular-sized models presented in Table~\ref{tab:comparison_sota}.

Across all groups, our ZIP variants achieve the best or highly competitive counting accuracy within their parameter-scale groups, while maintaining practical computational costs. These results validate the effectiveness and scalability of our ZIP framework, making it suitable for real-world deployments under various computational constraints.

\begin{figure}[t]
    \centering
    \subfloat[Input Image\label{fig:vis_compare_image}]{
        \includegraphics[width=0.47\linewidth]{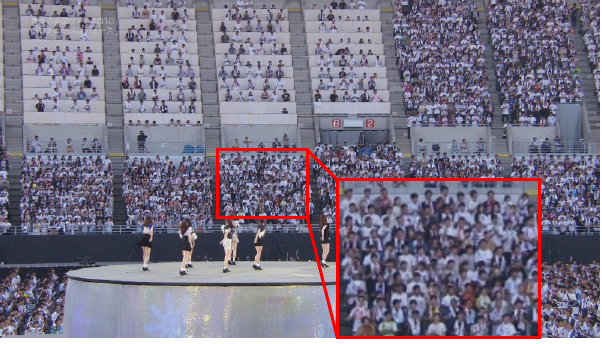}
    }
    \hfil
    \subfloat[GT Density Map\label{fig:vis_compare_gt_den}]{
        \includegraphics[width=0.47\linewidth]{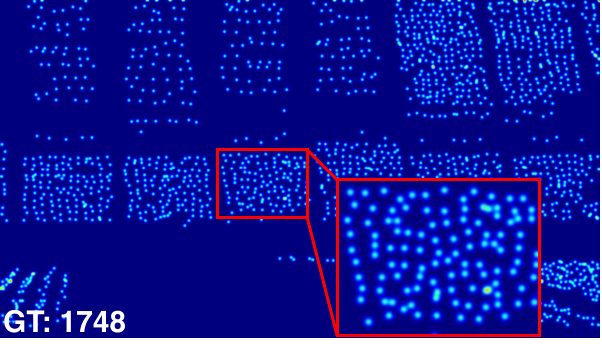}
    }
    \vfill
    \subfloat[Density Map (DMCount)\label{fig:vis_compare_dm_den}]{
        \includegraphics[width=0.47\linewidth]{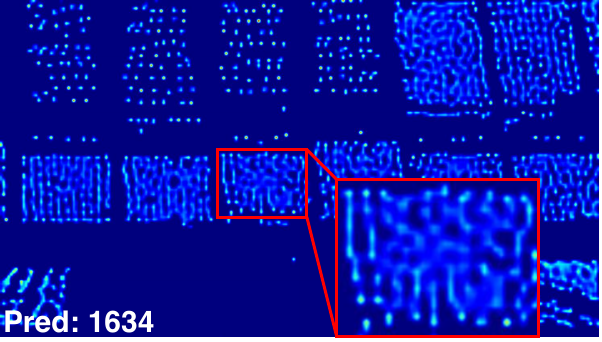}
    }
    \hfil
    \subfloat[Density Map (ZIP)\label{fig:vis_compare_ebc_zip_den}]{
        \includegraphics[width=0.47\linewidth]{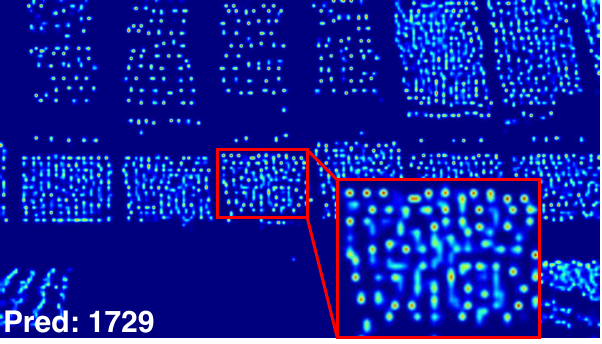}
    } 
    \caption{Qualitative comparison of DMCount and ZIP on an image from NWPU-Crowd (val). Both models utilize VGG19-based encoder-decoder structure as the backbone. 
    }
    \label{fig:vis_compare_1}
\end{figure}

\subsection{Overall Comparison of Frameworks}

We compare ZIP with several existing frameworks on the four benchmarks, shown in Table~\ref{tab:comparison_loss}. All methods are based on the same VGG19-based encoder-decoder backbone.
Among the compared VGG19-based frameworks, ZIP achieves the lowest errors on all four benchmarks. Notably, when compared directly with its EBC baseline, our ZIP model demonstrates significant performance gains across all datasets, reducing the MAE by a margin of 3.2\% to 14.3\%. These improvements validate the efficacy of zero-inflated Poisson regression for counting.

We also evaluate ZIP and DMCount on the same NWPU-Crowd validation image, using an identical VGG19-based backbone for fair comparison.
As shown in Fig.~\ref{fig:vis_compare_1}, the two density maps differ remarkably.
DMCount produces elongated, ribbon-like streaks inside each dense block, which is a typical artifact of its transport-based loss.
In contrast, ZIP yields compact, circular kernels that more closely resemble the point annotations, preserving the grid structure of the display.
As for accuracy, DMCount under-counts the scene by 114 heads (-6.5\%), while ZIP misses only 19 heads (-1.1\%), achieving a six-fold error reduction.

\begin{revblock}
\subsection{\minorrevreplace{Empirical Effect of Training Resolution}{Effect of Training Resolution on Accuracy and Training Time
}}
\label{sec:resolution_generalization}

\minorrevreplace{
The bound in Sec.~\ref{sec:risk_bound} suggests that the difference between local ZIP NLL supervision and OT-based supervision becomes more visible as the number of spatial blocks increases. 
We test this scaling behavior on UCF-QNRF by varying the training resolution while keeping the same VGG19-based encoder-decoder backbone, block size 32, and training protocol.
}{
Motivated by the different dependence on the number of spatial blocks in the bounds of Sec.~\ref{sec:risk_bound}, we compare ZIPNLL and DMCount (EBC) on UCF-QNRF across different training resolutions, using the same VGG19-based encoder-decoder backbone, block size 32, and training protocol.
}
The resulting grid size increases from $7\times7$ blocks at resolution $224\times224$ to $28\times28$ blocks at resolution $896\times896$.

\minorrevreplace{
Table~\ref{tab:res_block_size} reports counting accuracy and pure training time per epoch.
Compared with DMCount (EBC), ZIPNLL achieves nearly identical MAE at the lowest resolution and clear gains at medium-to-high resolutions, with the largest relative improvement of 5.1\% at training resolution 672.
This behavior is broadly consistent with the different spatial scaling of the two objectives discussed in Sec.~\ref{sec:risk_bound}.
ZIPNLL is also consistently faster per epoch, although the speedup decreases from $2.7\times$ to $1.4\times$ as the resolution increases, since its local likelihood avoids the heavier Sinkhorn-based OT computation.
}{
Table~\ref{tab:res_block_size} reports counting accuracy and pure training time per epoch.
ZIPNLL achieves relative MAE gains of 0.3\%, 3.1\%, 5.1\%, and 3.1\% as the training resolution increases from 224 to 896.
The gains are larger at medium-to-high resolutions than at 224, but are non-monotonic and peak at 672; thus, the empirical trend is broadly compatible with, but does not directly establish, the asymptotic scaling difference discussed in Sec.~\ref{sec:risk_bound}.
ZIPNLL is also consistently faster per epoch because its local likelihood avoids Sinkhorn-based OT, although the speedup decreases from $2.7\times$ to $1.4\times$ as the resolution increases.
}
\end{revblock}

\begin{table}[t]
\centering
\begin{revblock}
\caption{
Sensitivity to binning strategies on ShanghaiTech A and B using the same VGG19-based ZIP model.
Automatic strategies derive bin boundaries and centers only from training-set blockwise count histograms.
}
\label{tab:bin_sensitivity}
\resizebox{\columnwidth}{!}{%
\begin{tabular}{l|cc|c|cc}
\toprule
\multirow{2}{*}{\textbf{Bin Setting}} &
\multirow{2}{*}{\textbf{Rule-based}} &
\multirow{2}{*}{\textbf{Shared}} &
\multirow{2}{*}{\textbf{\#Bins}} &
\multicolumn{2}{c}{\textbf{MAE}} \\
& & & & \textbf{SHA} & \textbf{SHB} \\
\midrule
Default & \checkmark & -- & 14 / 10 & \textbf{58.7} & \textbf{6.0} \\
Direct Regression & N/A & N/A & N/A & 59.5 & 6.3 \\
Auto-Hist & \checkmark & -- & 14 / 11 & \underline{59.2} & \underline{6.1} \\
Coarse-Hist & \checkmark & -- & 6 / 6 & 60.1 & 6.5 \\
Fine-Hist & \checkmark & -- & 15 / 13 & 59.6 & 6.2 \\
Shared-16 & \checkmark & \checkmark & 14 / 14 & 59.4 & 6.3 \\
\bottomrule
\end{tabular}%
}
\end{revblock}
\end{table}

\begin{figure*}[!t]
    \centering
    \subfloat[Input Image\label{fig:xai_img}]{
        \includegraphics[width=0.31\textwidth]{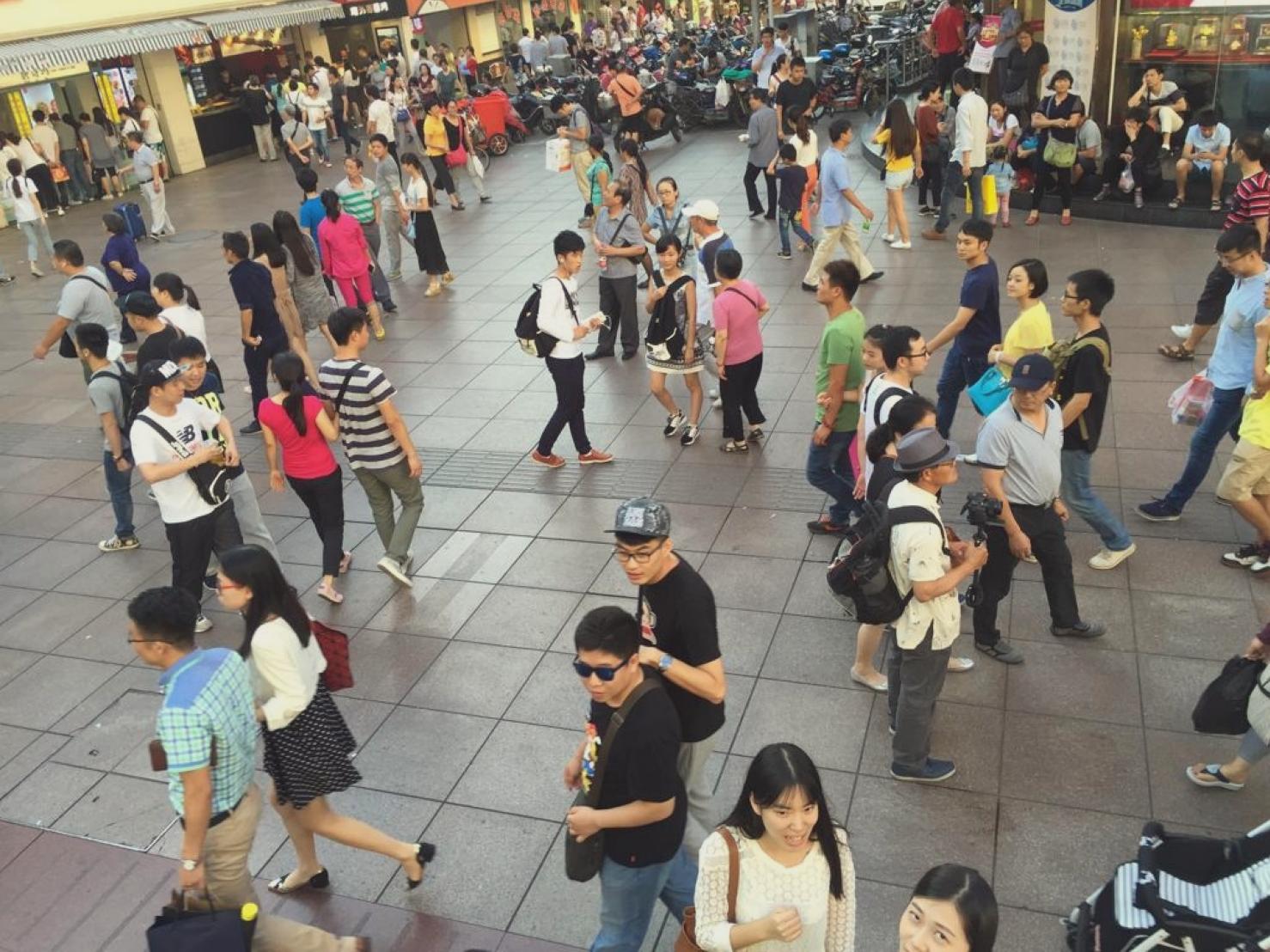}
    }
    \hfil
    \subfloat[Structural Zero Map $\boldsymbol{\pi}$\label{fig:xai_structural_zero}]{
        \includegraphics[width=0.31\textwidth]{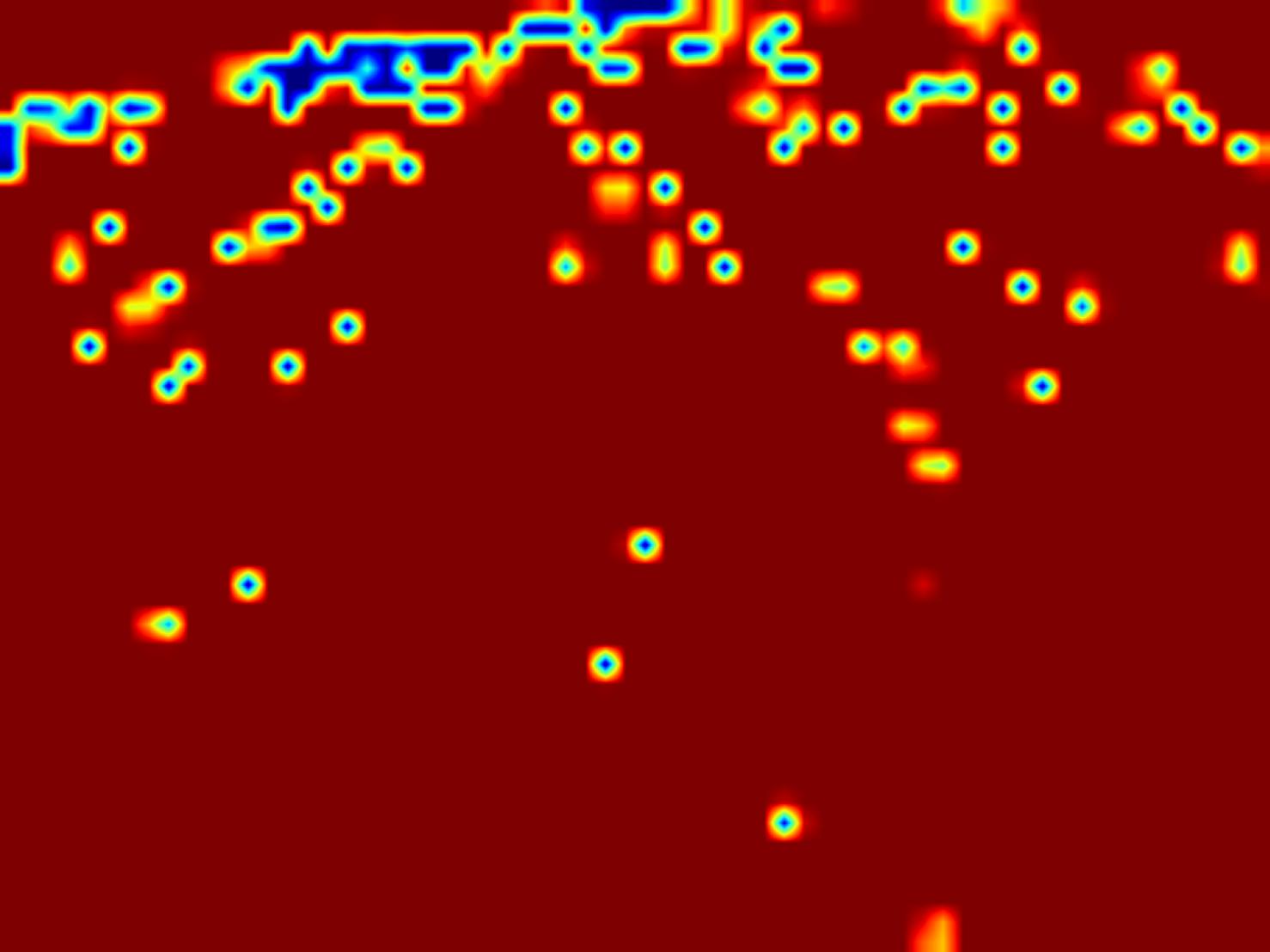}
    }    
    \hfil
    \subfloat[Lambda Map $\boldsymbol{\lambda}$\label{fig:xai_lambda}]{
        \includegraphics[width=0.31\textwidth]{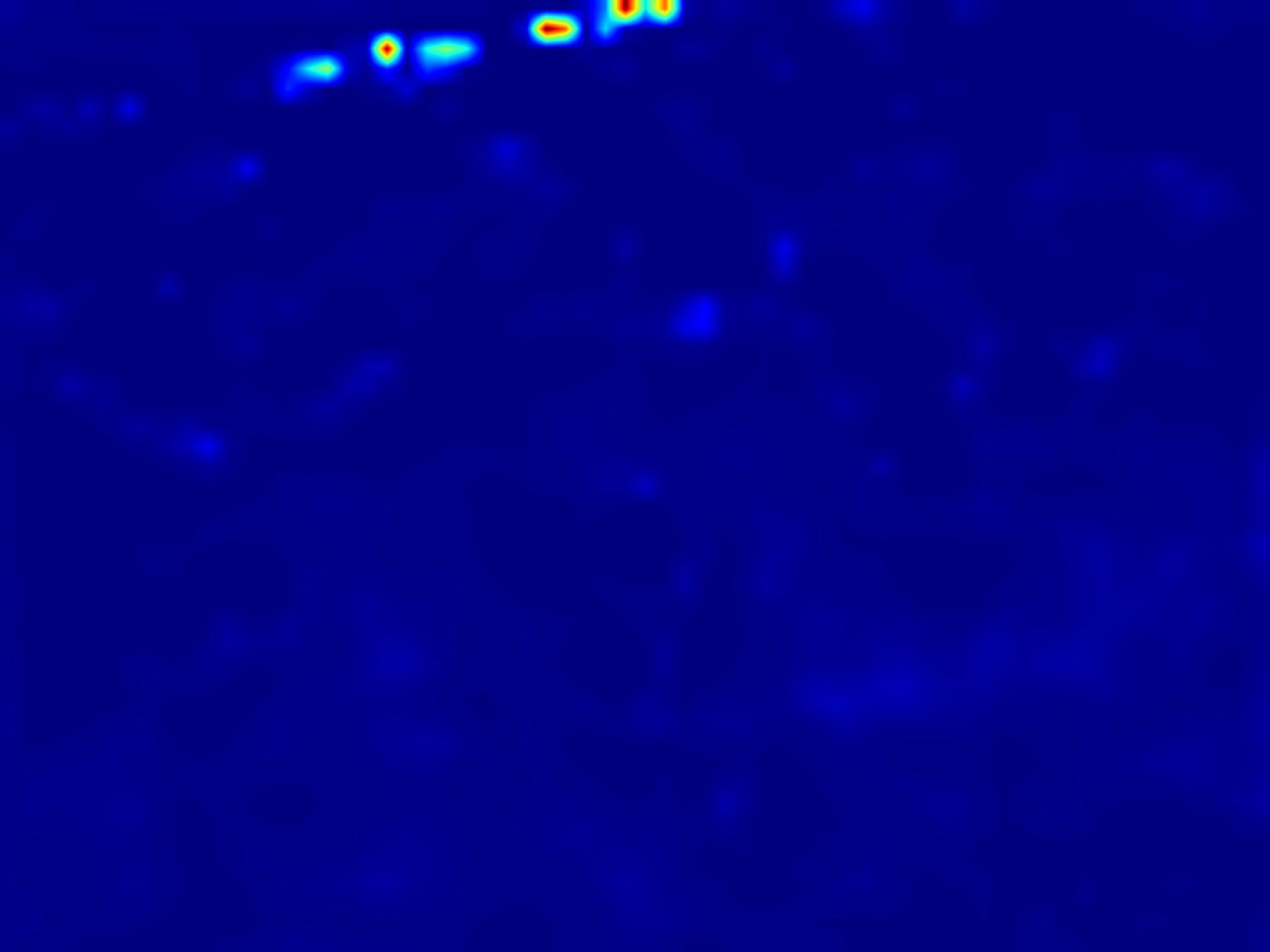}}
    \vfil
    \subfloat[Sampling Zero Map\label{fig:xai_sampling_zero}]{
        \includegraphics[width=0.31\textwidth]{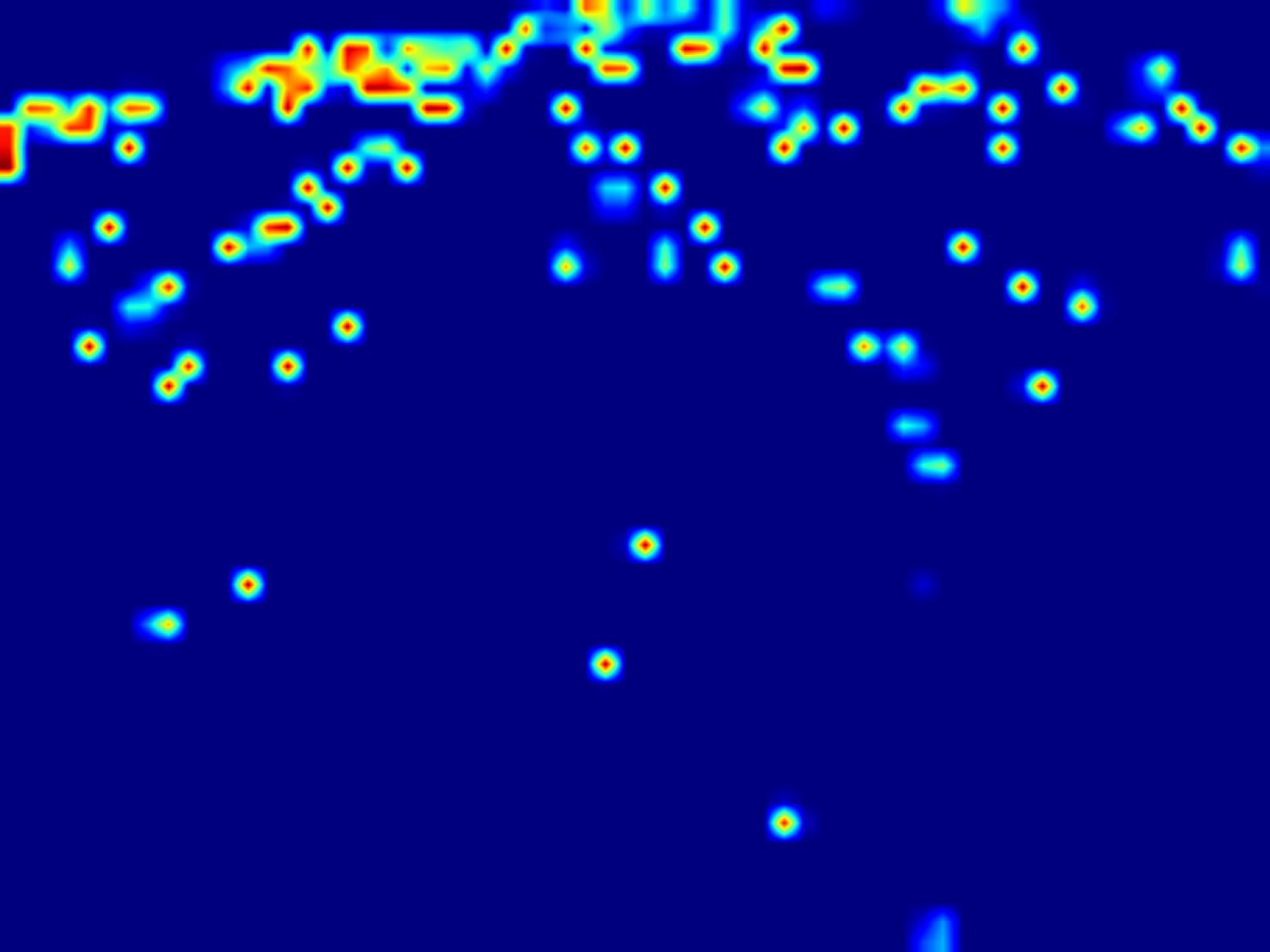}
    }
    \hfil
    \subfloat[Complete Zero Map\label{fig:xai_complete_zero}]{
        \includegraphics[width=0.31\textwidth]{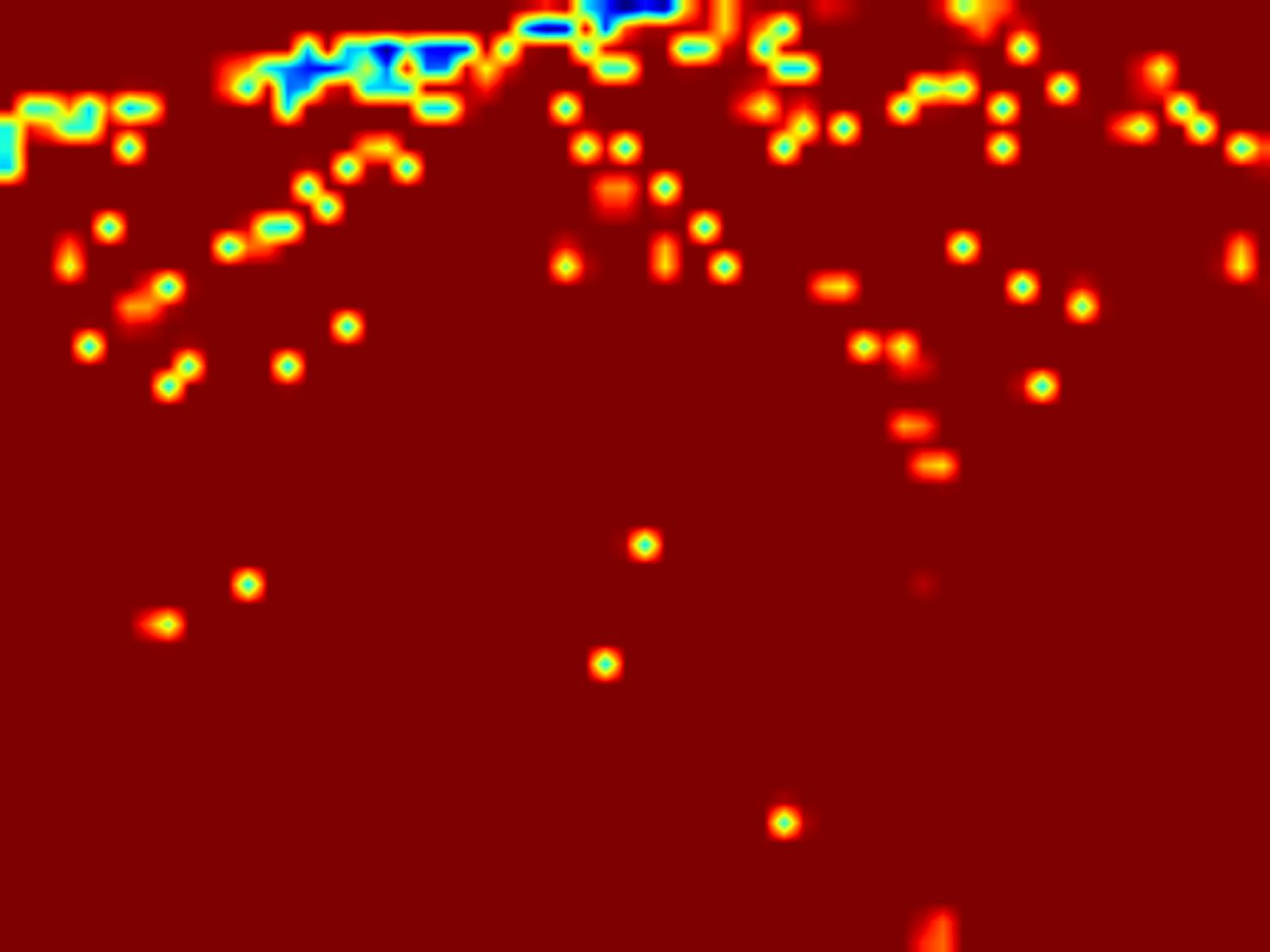}
    }
    \hfil
    \subfloat[Predicted Density Map\label{fig:xai_density}]{
        \includegraphics[width=0.31\textwidth]{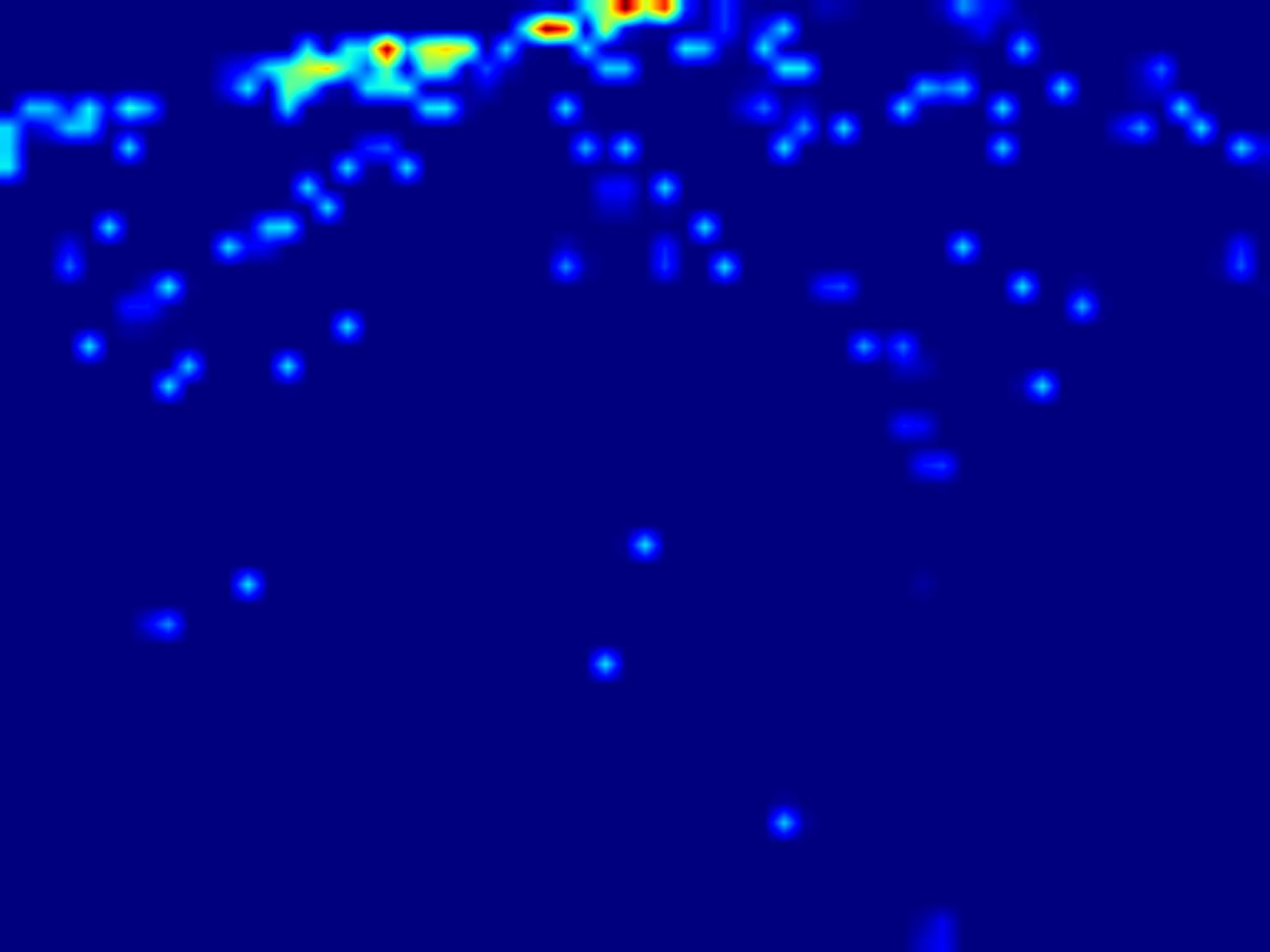}
    }
    \caption{
    \minorrevreplace{Visualization of the interpretability of our ZIP framework. This figure decomposes ZIP-B's prediction on a test image from ShanghaiTech B into its key components, illustrating how the model separates structurally irrelevant regions from informative ones.}{Visualization of ZIP's learned zero-inflation and Poisson-rate components on a ShanghaiTech B test image.}
    (a) Input Image: A representative test image from the ShanghaiTech B dataset.
    \minorrevreplace{
    (b) Structural Zero Map $\boldsymbol{\pi}$: Shows the model’s estimated probability that a block is structurally empty (\eg background, non-head regions). Red regions indicate high structural-zero probability, while blue/cyan spots highlight likely head-center regions, effectively masking irrelevant areas like pavement.
    }{
    (b) Structural Zero Map $\boldsymbol{\pi}$: Shows the learned zero-inflation map. Under the structural-zero interpretation, high values (red) occur mainly in non-head regions, while low values (blue/cyan) highlight candidate head-center regions.
    }
    (c) Lambda Map $\boldsymbol{\lambda}$: The predicted Poisson rate map, representing expected counts per block.
    \minorrevreplace{
    (d) Sampling Zero Map: Visualizes the probability of zero counts due to sampling error (\ie near-head-center blocks receive zero counts due to the point-based labels). Computed as $(1 - \boldsymbol{\pi}) \otimes \exp(-\boldsymbol{\lambda})$, it peaks at sparser head-center regions.
    }{
    (d) Sampling Zero Map: Shows $(1 - \boldsymbol{\pi}) \otimes \exp(-\boldsymbol{\lambda})$, the Poisson-component contribution to the probability of a zero count, which we interpret as a sampling-zero score.
    }
    \minorrevreplace{
    (e) Complete Zero Map: The total probability of a zero count from any source, combining both structural and sampling zeros.
    }{
    (e) Complete Zero Map: Shows the total predicted probability of a zero count,
    $\boldsymbol{\pi} + (1-\boldsymbol{\pi})\otimes\exp(-\boldsymbol{\lambda})$,
    combining the zero-inflation and Poisson-component contributions.
    }
    (f) Predicted Density Map $\boldsymbol{Y}^*$: The final predicted density map, computed as $(1 - \boldsymbol{\pi}) \otimes \boldsymbol{\lambda}$, integrating both spatial suppression and expected counts.
    \minorrevreplace{
    Together, these visualizations show how ZIP disentangles \emph{where} people are located from \emph{how densely} they are present.
    }{
    Together, these visualizations illustrate the complementary roles of the zero-inflation and Poisson-rate components in spatial gating and local count estimation.
    }
    }
    \label{fig:xai}
\end{figure*}

\begin{revblock}
\subsection{Bin Generation and Sensitivity}
\label{sec:bin_sensitivity}

The ZIP likelihood does not require binning in principle: the direct-regression variant in Table~\ref{tab:ablation_full} directly predicts non-negative rates without cross-entropy supervision.
Our final model uses an EBC-style bin-expectation head only to stabilize long-tailed positive-count estimation.
The default bins are built from the training-set blockwise count histogram with a frequency-aware rule: frequent low counts are kept as singleton bins, rare high counts are merged into wider intervals, and the last bin is open-ended; bin centers are training-set empirical means.
Detailed boundaries and cut points are provided in the supplement.

To assess sensitivity to discretization, we compare the default bins with four histogram-driven alternatives on ShanghaiTech A/B: Auto-Hist, Coarse-Hist, Fine-Hist, and Shared-16.
All variants use the same backbone, training schedule, block size, and loss weights.
Table~\ref{tab:bin_sensitivity} shows that the default bins achieve the best MAE, but the alternatives remain close.
For example, Auto-Hist is only 0.5 MAE (0.9\%) worse on ShanghaiTech A and 0.1 MAE (1.7\%) worse on ShanghaiTech B, while Shared-16 remains competitive despite using one shared bin configuration.
Together with the bin-free direct-regression result, these findings indicate that ZIP is not tied to fragile dataset-specific bin engineering; it can be used without bins, and the final bin-expectation implementation is robust to reasonable histogram-driven choices.
\end{revblock}

\subsection{Model Interpretation}
\label{sec:model_interpretation}

A key advantage of the proposed ZIP framework is its inherent interpretability.
By decomposing the counting process into distinct probabilistic components, ZIP enables us to diagnose the model's behavior by separately analyzing where it focuses and how much it counts, as illustrated in Fig.~\ref{fig:xai}.

\minorrevreplace{The structural zero map ($\boldsymbol{\pi}$) functions as a learned attention or segmentation-like mechanism. This branch estimates the probability that a given block is structurally irrelevant to crowd counting (\eg background, buildings, torso, or peripheral head parts) and should be suppressed.}{The $\boldsymbol{\pi}$ branch produces a zero-inflation probability map that acts as a soft spatial gate. Under the structural-zero interpretation, high values correspond to blocks likely to be suppressed (\eg background, buildings, torso, or peripheral head parts).}
As shown in Fig.~\ref{fig:xai_structural_zero}, the model assigns high \minorrevreplace{structural-zero probability}{zero-inflation probability} (deep red) to background areas like pavement, while assigning low probability (blue/cyan) to compact regions containing head centers.
\minorrevreplace{Note that this spatial disentanglement is learned without any segmentation supervision, showing that ZIP can implicitly separate relevant and irrelevant regions using only point-level annotations.}{This spatial selectivity emerges without segmentation supervision and is quantified below.}

In parallel, the lambda map $\boldsymbol{\lambda}$ captures the expected count per block. As visualized in Fig.~\ref{fig:xai_lambda}, high values appear in the upper part of the image where the crowd is denser due to perspective.


\begin{table*}[t]
\centering
\begin{revblock}
\caption{\minorrevreplace{Quantitative validation of zero-map semantics}{Quantitative analysis of learned zero-map behavior} on in-domain evaluation splits. $\boldsymbol{\pi}$-Struct AUC evaluates whether $\boldsymbol{\pi}$ ranks far-background \minorrevreplace{structural-zero blocks}{zero blocks used as structural-zero proxies} above head-neighborhood blocks. Sampling-Zero AP evaluates the sampling-zero score $(1-\boldsymbol{\pi})\exp(-\boldsymbol{\lambda})$ on zero-count blocks. $\boldsymbol{\pi}$-Retention AP evaluates head-containing block detection using $1-\boldsymbol{\pi}$. $\boldsymbol{\pi}$-BGL and $\boldsymbol{\pi}-$HSR denote background leakage and head suppression rates at threshold 0.5.}
\label{tab:zero_map_semantics}
\begin{tabular}{l|ccccc}
\toprule
\textbf{Dataset} & $\boldsymbol{\pi}$-\textbf{Struct AUC} (\%) $\uparrow$ & \textbf{Sampling-Zero AP} (\%) $\uparrow$ & $\boldsymbol{\pi}$-\textbf{Retention AP} (\%) $\uparrow$ & $\boldsymbol{\pi}$-\textbf{BGL}@0.5  (\%) $\downarrow$ & $\boldsymbol{\pi}-$\textbf{HSR}@0.5 (\%) $\downarrow$ \\
\midrule
ShanghaiTech A        & 91.9 & 83.7 & 75.7 & 0.2 & 36.4 \\
ShanghaiTech B        & 90.9 & 74.2 & 81.6 & 0.0 & 27.9 \\
UCF-QNRF              & 92.3 & 80.3 & 87.1 & 0.1 & 19.4 \\
NWPU-Crowd (Val)      & 95.8 & 68.1 & 72.5 & 0.0 & 39.2 \\
\bottomrule
\end{tabular}%
\end{revblock}
\end{table*}

\minorrevreplace{The sampling zero map (Fig.~\ref{fig:xai_sampling_zero}) models the probability that a block receives a zero count due to sampling effects (\ie head-center blocks receive a zero count because of the point-based annotation).}{The map $(1-\boldsymbol{\pi})\exp(-\boldsymbol{\lambda})$ in Fig.~\ref{fig:xai_sampling_zero} is the Poisson-component contribution to the zero probability and is interpreted as a sampling-zero score.}
\minorrevreplace{
It is computed as $(1 - \boldsymbol{\pi}) \otimes \exp(-\lambda)$, and is highest where the model is confident a person is present (low $\boldsymbol{\pi}$), but the predicted count is modest (moderate $\lambda$).
}{
It becomes large when both the retention probability $(1-\boldsymbol{\pi}$ and the Poisson zero probability $\exp(-\boldsymbol{\lambda})$ are high, corresponding to low $\boldsymbol{\pi}$ and relatively small $\boldsymbol{\lambda})$.
}

\minorrevreplace{The complete zero probability map (Fig.~\ref{fig:xai_complete_zero}) represents the overall likelihood that a block receives a zero count from any source (either structural or sampling).}{The complete zero probability map in Fig.~\ref{fig:xai_complete_zero} combines the zero-inflation and Poisson-component contributions.} It is defined as $\boldsymbol{\pi} + (1- \boldsymbol{\pi})\otimes \exp(-\lambda)$.
This map is often dominated by the \minorrevreplace{structural zero}{zero-inflation} component, as the background makes up the majority of the image. It can be viewed as an inverse attention map, where non-red regions highlight areas the model expects to contain people.

Finally, the predicted density map $\boldsymbol{Y}^*$ (Fig.~\ref{fig:xai_density}) is computed as $(1 - \boldsymbol{\pi}) \otimes \boldsymbol{\lambda}$, combining spatial suppression and count estimation. This yields a clean and well-localized density prediction, with high activations concentrated near annotated head centers.

Together, these components demonstrate how ZIP not only improves counting accuracy, but also offers fine-grained interpretability by isolating both spatial relevance and count magnitude across the image.

\begin{revblock}
\noindent\textbf{\minorrevreplace{Quantitative validation of zero-map semantics.}{Quantitative analysis of zero-map behavior.}}
The qualitative decomposition above suggests that $\boldsymbol{\pi}$ behaves as a structural-zero map, but structural and sampling zeros are latent variables and are not separately annotated in crowd-counting datasets.
We therefore construct evaluation-only pseudo labels from point annotations to quantify this behavior.
For each output block $b$, let $Y_b$ denote the integer count obtained by assigning raw head points to the model's output grid.
We define head-containing blocks as
\begin{equation}
    \mathcal{H}=\{b:Y_b>0\},
\end{equation}
near-head zero blocks as
\begin{equation}
    \mathcal{S}=\{b:Y_b=0,\ d(b,\mathcal{P})\le B\},
\end{equation}
and structural-zero proxy blocks as
\begin{equation}
    \mathcal{Z}=\{b:Y_b=0,\ d(b,\mathcal{P})>B\},
\end{equation}
where $\mathcal{P}$ is the set of annotated head points, $d(b,\mathcal{P})$ is the minimum distance from block $b$ to any head point, and $B$ is the dataset-specific block size.
These pseudo labels are used only for evaluation and never for training.

Table~\ref{tab:zero_map_semantics} reports three threshold-free ranking metrics and two $\boldsymbol{\pi}$-specific operating-point diagnostics.
Let
\begin{align}
    s_{\mathrm{struct}}(b) &= \pi_b, \notag\\
    s_{\mathrm{samp}}(b)   &= (1-\pi_b)\exp(-\lambda_b), \notag \\
    s_{\mathrm{ret}}(b)    &= 1-\pi_b .
\end{align}
The $\boldsymbol{\pi}$-Struct AUC metric is defined as
\begin{equation}
    \mathrm{AUC}\big(s_{\mathrm{struct}};\mathcal{Z},\mathcal{H}\cup\mathcal{S}\big),
\end{equation}
and measures whether the structural-zero score $\pi_b$ ranks far-background \minorrevreplace{structural-zero blocks}{structural-zero proxy blocks} above head-containing and near-head blocks.
The ideal value is high.

Sampling-Zero AP is defined as
\begin{equation}
    \mathrm{AP}\big(s_{\mathrm{samp}};\mathcal{S},\mathcal{Z}\big),
\end{equation}
and evaluates whether the sampling-zero score $(1-\pi_b)\exp(-\lambda_b)$ ranks near-head zero blocks above far-background zero blocks among zero-count regions.
We use AP rather than AUROC because near-head sampling zeros form a relatively rare subset among zero-count blocks; AP better reflects whether high sampling-zero scores concentrate on these rare near-head zero blocks.

The $\boldsymbol{\pi}$-Retention AP metric is defined as
\begin{equation}
    \mathrm{AP}\big(s_{\mathrm{ret}};\mathcal{H},\mathcal{S}\cup\mathcal{Z}\big),
\end{equation}
and measures whether the retention score $1-\pi_b$ ranks positive-count blocks above zero-count blocks.
It evaluates whether the structural-zero branch preserves count-relevant blocks instead of suppressing them.
\end{revblock}

\begin{revblock}
The $\boldsymbol{\pi}$ Background Leakage metric is defined as
\begin{equation}
    \boldsymbol{\pi}\text{-}\mathrm{BGL@0.5}
    =
    \frac{1}{|\mathcal{Z}|}
    \sum_{b\in\mathcal{Z}}
    \mathbb{I}[1-\pi_b\ge 0.5],
\end{equation}
which measures the fraction of \minorrevreplace{structural-zero blocks}{structural-zero proxy blocks} that would be incorrectly retained under a hard threshold.
The $\boldsymbol{\pi}$ Head Suppression Rate is defined as
\begin{equation}
    \boldsymbol{\pi}\text{-}\mathrm{HSR@0.5}
    =
    \frac{1}{|\mathcal{H}|}
    \sum_{b\in\mathcal{H}}
    \mathbb{I}[1-\pi_b<0.5],
\end{equation}
which measures the fraction of positive-count blocks that would be suppressed under the same hard threshold.
\end{revblock}

\begin{revblock}
Since structural and sampling zeros are latent and the sets
$\mathcal{H}$, $\mathcal{S}$, and $\mathcal{Z}$ are point-derived
proxy labels, Table~\ref{tab:zero_map_semantics} should be
interpreted as a diagnostic of whether the learned maps are
consistent with the intended ZIP decomposition, rather than as
direct ground-truth validation of the latent variables.
Under this diagnostic, $\boldsymbol{\pi}$ achieves consistently
high $\boldsymbol{\pi}$-Struct AUC on all four benchmarks, ranging
from 90.9\% to 95.8\%.
This suggests that the structural-zero score provides a useful
threshold-free ranking of far-background zero blocks above
head-containing and near-head blocks, despite receiving no
segmentation supervision.
However, this ranking result does not imply that
$\boldsymbol{\pi}$ is a calibrated binary segmentation mask.

\begin{table}[t]
\centering
\begin{revblock}
\caption{
Cross-dataset evaluation of the structural-zero branch from NWPU-Crowd to ShanghaiTech A/B.
ZIP-B is trained on NWPU-Crowd and evaluated on target datasets without target fine-tuning.
Metrics are computed on the predicted $\boldsymbol{\pi}$ maps using the same proxy labels as in Table~\ref{tab:zero_map_semantics}.
}
\label{tab:zero_map_cross_dataset}
\begin{tabular}{l|cc}
\toprule
\textbf{Metric} & \textbf{NWPU $\rightarrow$ SHA} & \textbf{NWPU $\rightarrow$ SHB} \\
\midrule
$\boldsymbol{\pi}$-\textbf{Struct AUC} (\%) $\uparrow$     & 92.0 & 93.1 \\
$\boldsymbol{\pi}$-\textbf{Retention AP} (\%) $\uparrow$   & 71.3 & 75.0 \\
$\boldsymbol{\pi}$-\textbf{BGL}@0.5 (\%) $\downarrow$      & 0.1  & 0.0  \\
$\boldsymbol{\pi}$-\textbf{HSR}@0.5 (\%) $\downarrow$      & 35.5 & 36.2 \\
\bottomrule
\end{tabular}%
\end{revblock}
\end{table}

Sampling-Zero AP is also non-trivial across datasets.
Since this score is computed as
$(1-\boldsymbol{\pi})\exp(-\boldsymbol{\lambda})$, it evaluates the
joint $\boldsymbol{\pi}$--$\boldsymbol{\lambda}$ decomposition rather
than the structural-zero branch alone.
We therefore interpret it as evidence that near-head zero blocks
tend to receive higher sampling-zero scores than far-background
zero blocks, not as direct proof that the unobserved latent causes
are perfectly disentangled.
The $\boldsymbol{\pi}$-Retention AP values are moderate to high,
indicating that $1-\boldsymbol{\pi}$ generally ranks positive-count
blocks above zero-count blocks.
At the same time, the lower retention values on ShanghaiTech A
and NWPU-Crowd show that positive-region preservation is not
perfect.

The operating-point diagnostics further reveal a
specificity--retention trade-off.
The near-zero $\boldsymbol{\pi}$-BGL@0.5 values show that a
threshold of 0.5 rarely retains far-background blocks.
However, the non-negligible $\boldsymbol{\pi}$-HSR@0.5 values show
that the same threshold would also suppress a noticeable fraction
of positive-count blocks.
Thus, $\boldsymbol{\pi}$ should be viewed as a soft structural-zero
gating signal rather than a hard foreground/background mask.
At test time, ZIP never binarizes $\boldsymbol{\pi}$; instead, it
uses the continuous expectation
$(1-\boldsymbol{\pi})\otimes\boldsymbol{\lambda}$ as the predicted
density map.
A block counted as suppressed by $\boldsymbol{\pi}$-HSR@0.5 can
therefore still contribute to the final density map in proportion
to its continuous retention weight and predicted Poisson rate.
However, this soft coupling does not fully eliminate false
negatives: if both $1-\boldsymbol{\pi}$ and
$\boldsymbol{\lambda}$ are underestimated in visually challenging
regions, under-counting can still occur.
Overall, the zero-map diagnostics support that
$\boldsymbol{\pi}$ learns a meaningful soft structural-zero ranking,
while also exposing hard-threshold calibration and false-negative
risks.
\end{revblock}

\begin{revblock}
\subsection{Cross-Dataset Evaluation from NWPU-Crowd}
\label{sec:cross_dataset_zero}

\minorrevreplace{The in-domain analysis above verifies the zero-map behavior under the training distribution.}{The preceding analysis characterizes the learned zero-map behavior on in-domain evaluation splits.}
To test whether the structural-zero branch merely memorizes source-domain background layouts, we conduct a source-only transfer evaluation from NWPU-Crowd to ShanghaiTech A/B.
NWPU-Crowd is used as the source because it is the largest and most diverse benchmark in our evaluation.
ZIP-B is trained only on NWPU-Crowd and evaluated directly on ShanghaiTech A/B; no target images, target annotations, or target fine-tuning are used during training or model selection, and target annotations are used only for final evaluation.

\begin{table}[t]
\centering
\begin{revblock}
\caption{
Reported cross-dataset counting performance from NWPU-Crowd to ShanghaiTech A/B.
ZIP-B is trained on NWPU-Crowd and evaluated directly on the target datasets without target-domain adaptation. Prior results are reported from the literature and may use different architectures, training protocols, or target-domain information.
}
\label{tab:cross_dataset_counting_nwpu}
\begin{tabular}{l|c|cc|cc}
\toprule
\multirow{2}{*}{\textbf{Method}} & \multirow{2}{*}{\textbf{DA?}} &
\multicolumn{2}{c|}{\textbf{NWPU $\rightarrow$ SHA}} &
\multicolumn{2}{c}{\textbf{NWPU $\rightarrow$ SHB}} \\
& & \textbf{MAE} $\downarrow$ & \textbf{RMSE} $\downarrow$
& \textbf{MAE} $\downarrow$ & \textbf{RMSE} $\downarrow$ \\
\midrule
DAOT \cite{zhu2023daot} & \checkmark              & 79.9 & 135.0 & \underline{8.8} & \textbf{14.8} \\
FSIM \cite{zhu2023find} & \checkmark              & 81.0 & 133.2 & 9.1  & 16.0 \\
MGOT \cite{zhong2025multi} & \checkmark              & \underline{69.7} & \underline{128.7} & 9.4 & 15.3 \\
\midrule
\textbf{ZIP-B (ours)} & -- & \textbf{59.4} & \textbf{100.6} & \textbf{8.6} & \underline{14.9} \\
\bottomrule
\end{tabular}%
\end{revblock}
\end{table}

Table~\ref{tab:zero_map_cross_dataset} evaluates the transferred $\boldsymbol{\pi}$ branch using the same $\boldsymbol{\pi}$-based diagnostics as Table~\ref{tab:zero_map_semantics}.
Sampling-Zero AP is omitted because it depends on both $\boldsymbol{\pi}$ and $\boldsymbol{\lambda}$.
The $\boldsymbol{\pi}$-Struct AUC values remain high on both targets, reaching 92.0\% on ShanghaiTech A and 93.1\% on ShanghaiTech B, indicating that $\boldsymbol{\pi}$ still ranks far-background zero blocks above head-neighborhood blocks under target-domain shift.
The $\boldsymbol{\pi}$-Retention AP values are 71.3\% and 75.0\%, while background leakage remains close to zero.
The non-negligible $\boldsymbol{\pi}$-HSR@0.5 values should be interpreted as conservative hard-threshold diagnostics, since ZIP uses the continuous expectation $(1-\boldsymbol{\pi})\otimes\boldsymbol{\lambda}$ rather than binarizing $\boldsymbol{\pi}$.

Table~\ref{tab:cross_dataset_counting_nwpu} further reports source-only counting performance in the same transfer setting.
ZIP-B obtains 59.4/100.6 MAE/RMSE on ShanghaiTech A and 8.6/14.9 on ShanghaiTech B.
Compared with reported cross-dataset methods, ZIP-B shows strong transfer performance even without target-domain adaptation.
Since these methods may use different architectures and protocols, this table is intended as contextual evidence rather than a controlled domain-adaptation benchmark.
Overall, the results suggest that a ZIP model trained on a sufficiently diverse source dataset can transfer meaningful structural-zero ranking and usable count estimates to smaller target benchmarks.
\end{revblock}

\begin{table*}[t]
\caption{Backbone scalability of ZIP variants on ShanghaiTech A \& B and UCF-QNRF. All variants use the same ZIP head and loss formulation while varying backbone capacity from 0.81M to 105.60M parameters. Best results are shown in \textbf{bold} and the second best are \underline{underlined}.}
\label{tab:comparison_all}
\centering
\begin{tabular}{c|c|ccccccccc}
\toprule
\multirow{2}{*}{\textbf{Suffix}} & \multirow{2}{*}{\textbf{Size (M)}} & \multicolumn{3}{c}{\textbf{ShanghaiTech A}}        & \multicolumn{3}{c}{\textbf{ShanghaiTech B}}      & \multicolumn{3}{c}{\textbf{UCF-QNRF}}               \\
                                  &                                    & \textbf{MAE}   & \textbf{RMSE}  & \textbf{NAE (\%)} & \textbf{MAE}  & \textbf{RMSE} & \textbf{NAE (\%)} & \textbf{MAE}   & \textbf{RMSE}   & \textbf{NAE (\%)} \\ \midrule
-P                         & 0.8                               & 71.1          & 109.6         & 16.6            & 8.2          & 12.6         & 6.9             & 96.2          & 161.8          & 14.4            \\
-N                         & 3.3                               & 58.8          & 94.6          & 14.1            & 7.7          & 12.1         & 6.3             & 86.4          & 147.6          & 12.6            \\
-T                         & 10.6                              & 56.3          & 90.5    & 13.2            & 6.6          & 9.9          & 5.5             & 76.0          & 129.4          & 11.1            \\
-S                         & 33.7                              & \underline{55.1}    & \underline{88.9}          & \underline{11.9}      & \underline{5.8}    & \underline{9.2}    & \underline{4.5}       & \underline{73.3}    & \underline{125.0}    & \underline{10.4}      \\
-B                         & 105.6                             & \textbf{47.8} & \textbf{75.0} & \textbf{11.0}   & \textbf{5.5} & \textbf{8.6} & \textbf{4.4}    & \textbf{69.4} & \textbf{121.8} & \textbf{10.1}   \\ \bottomrule
\end{tabular}
\end{table*}

\subsection{Backbone Scalability of ZIP Variants}

Table~\ref{tab:comparison_all} reports the full results of the five ZIP variants---Pico (-P), Nano (-N), Tiny (-T), Small (-S), and Base (-B)---on ShanghaiTech A, ShanghaiTech B, and UCF-QNRF.
This comparison is designed to test whether ZIP scales with backbone capacity while keeping the counting head, loss formulation, and training protocol fixed.
%
The variants differ only in backbone capacity (from 0.8 M to 105.6 M parameters); all heads, losses and training settings are identical. From this table, we can observe three trends:

\begin{itemize}
    \item \textbf{Monotonic accuracy gain with scale.} From -P to -B every step produces lower error on all datasets. MAE on ShanghaiTech A drops from 71.1 to 47.8 (-32.8\%), while UCF-QNRF MAE falls from 96.2 to 69.4 (-27.9\%). A similar monotone decrease is observed for RMSE and NAE.
    \item \textbf{Diminishing returns.} The relative improvement per additional million parameters shrinks after the -T model (10 M). For example, moving from -T to -S cuts ShanghaiTech A MAE by 2.1\% whereas the jump from -N to -T yields 2.0$\times$ that reduction with fewer added parameters. This suggests that -T or -S may offer the best accuracy--efficiency trade-off for many deployments.
    \item \textbf{Consistent generalization across datasets.} Ranking of variants is identical on ShanghaiTech A/B and UCF-QNRF, indicating that gains are not dataset-specific but stem from the scalable ZIP formulation itself.
\end{itemize}
Overall, the results demonstrate that ZIP scales gracefully: the smallest model remains competitive for edge devices, whereas the largest model achieves state-of-the-art accuracy.

\begin{table}[t]
\centering
\begin{revblock}
\caption{
Compatibility with Mamba-based visual backbones on ShanghaiTech A \& B and UCF-QNRF.
EBC and ZIP variants use the same backbone within each pair.
}
\label{tab:mamba_backbones}
\resizebox{\columnwidth}{!}{%
\begin{tabular}{l|cc|cc|cc}
\toprule
\multirow{2}{*}{\textbf{Method}} &
\multicolumn{2}{c|}{\textbf{SHA}} &
\multicolumn{2}{c|}{\textbf{SHB}} &
\multicolumn{2}{c}{\textbf{UCF-QNRF}} \\
& \textbf{MAE} & \textbf{RMSE} & \textbf{MAE} & \textbf{RMSE} & \textbf{MAE} & \textbf{RMSE} \\
\midrule
VMambaCC & \textbf{51.9} & \textbf{81.3} & 7.5 & 12.5 & 88.4 & 144.7 \\
EBC + VMamba & 55.9 & 91.4 & 6.5 & 10.8 & 79.7 & 140.6 \\
ZIP + VMamba & 53.7 & 87.6 & 6.3 & 10.5 & 78.4 & 137.8 \\
\midrule
EBC + MambaVision & 53.1 & 88.6 & \underline{6.2} & \underline{10.1} & \underline{76.6} & \underline{134.1} \\
ZIP + MambaVision & \underline{52.0} & \underline{82.5} & \textbf{5.8} & \textbf{9.6} & \textbf{74.3} & \textbf{129.9} \\
\bottomrule
\end{tabular}%
}
\end{revblock}
\end{table}

\subsection{Compatibility with Mamba-Based Backbones}
\label{sec:mamba_backbones}

\begin{revblock}
To further examine whether ZIP is tied to CNN-style feature extractors, we evaluate it with two contemporary Mamba-style visual backbones: VMamba~\cite{liu2024vmamba} and MambaVision~\cite{hatamizadeh2025mambavision}. For each backbone, the EBC and ZIP variants in Table~\ref{tab:mamba_backbones} share the same architecture and training protocol, with only the counting framework changed. We also include the VMambaCC results reported in~\cite{ma2024vmambacc} as a reference for dedicated VMamba-based crowd counting.
\end{revblock}

\begin{revblock}
As shown in Table~\ref{tab:mamba_backbones}, ZIP remains effective when combined with Mamba-style visual backbones. With VMamba, ZIP reduces MAE over EBC on all three datasets and lowers RMSE on all three datasets, yielding maximum relative reductions of 3.9\% and 4.2\% in MAE and RMSE, respectively. These improvements suggest that zero-inflated modeling can provide complementary benefits to Mamba-style visual representations. The gains are more pronounced with MambaVision. ZIP reduces MAE/RMSE from 53.1/88.6 to 52.0/82.5 on ShanghaiTech A, from 6.2/10.1 to 5.8/9.6 on ShanghaiTech B, and from 76.6/134.1 to 74.3/129.9 on UCF-QNRF, achieving up to 6.5\% relative MAE reduction over EBC. The reported VMambaCC entries serve as an external reference, while the controlled comparison in this table is the paired EBC-versus-ZIP evaluation under the same backbone. Overall, these results indicate that ZIP is also compatible with modern Mamba-style visual backbones.
\end{revblock}

\subsection{Inference Speed}

To assess the practical deployment potential of our lightweight models, we measure the average inference speed of each ZIP variant across four representative hardware platforms: Apple M1 Pro CPU, Apple M1 Pro 14-core GPU, AMD Ryzen 9 5900X CPU (12-core), and NVIDIA RTX 3090 GPU.
All experiments are conducted on the ShanghaiTech B dataset using full-precision inference at an input resolution of $1024 \times 768$, which reflects real-world surveillance camera scenarios.
Inference speed is computed as the reciprocal of the average per-image inference time, where the time is measured individually for each test image.
The PyTorch version is 2.7.1.
Table~\ref{tab:comparison_speed} reports the average frames per second (FPS) for each model. Our smallest variant, ZIP-P, achieves the highest throughput on all platforms, reaching 122.2 FPS on AMD 5900X and over 1,000 FPS on RTX 3090, making it ideal for resource-constrained or real-time applications.
On the other end, ZIP-S delivers the best accuracy (MAE = 5.8) but at a higher computational cost, with CPU-side inference limited to under 5 FPS, making it suitable for accuracy-critical but latency-tolerant use cases.
ZIP-N and ZIP-T offer compelling accuracy-efficiency trade-offs. For example, ZIP-T achieves 18.3 FPS on AMD 5900X and 249.7 FPS on RTX 3090, while maintaining a competitive MAE of 6.6. These results indicate that mid-sized variants can enable near real-time performance even on high-end CPUs without requiring dedicated GPUs.
Overall, our ZIP family provides a scalable solution space, allowing deployment under varying hardware and latency constraints without compromising on count accuracy.

\begin{table}[t]
\caption{Inference speed (frames per second, FPS) of the lightweight  variants on ShanghaiTech B ($1024 \times 768$ resolution). Results are reported on four hardware platforms.}
\label{tab:comparison_speed}
\centering
\resizebox{\columnwidth}{!}{%
\begin{tabular}{c|c|cccc}
\toprule
\textbf{Suffix} & \textbf{MAE} & \textbf{M1 Pro CPU} & \textbf{M1 Pro GPU} & \textbf{5900X} & \textbf{3090} \\ \midrule
-P        & 8.2         & 9.9                                    & 118.6                                  & 122.2                                 & 1,012.0                                     \\
-N        & 7.7         & 5.2                                    & 63.1                                   & 50.6                                 & 548.1                                       \\
-T        & 6.6         & 1.8                                   & 27.7                                   & 18.3                                  & 249.7                                       \\
-S        & 5.8         & 0.4                                    & 6.8                                    & 4.6 
                     & 50.9 \\
\bottomrule
\end{tabular}
}
\end{table}

\begin{figure*}[t]
    \centering
    \subfloat[Background-only instance\label{fig:more_vis_background}]{
        \includegraphics[width=\textwidth]{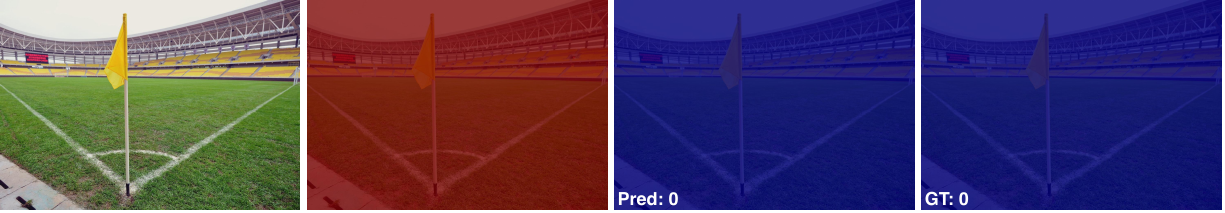}
    }
    \vfil
    \subfloat[Sparse scenario\label{fig:more_vis_sparse}]{
        \includegraphics[width=\textwidth]{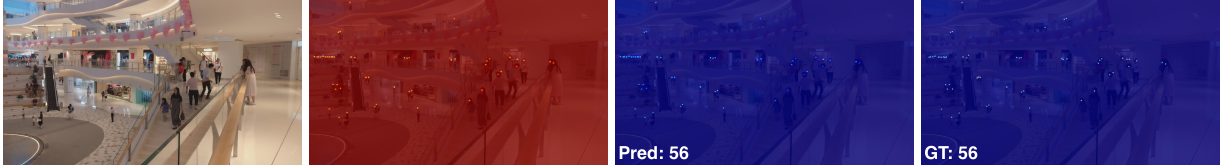}
    }
    \vfil
    \subfloat[Highly crowded scene\label{fig:more_vis_crowded}]{
        \includegraphics[width=\textwidth]{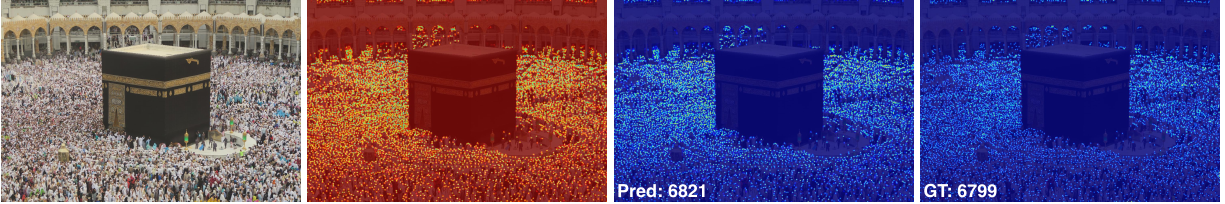}
    }
    \caption{
        Visualized results on background-only, sparse and highly congested scenes in NWPU-Crowd (predictions by ZIP-B). 
        Each row shows, from left to right, the input image, the predicted structural-zero map, the predicted density map, and the ground-truth density map. 
        In the background-only example (a), the zero-inflation branch assigns high probability across the image, yielding an effectively zero density map.
        In the sparse case (b), ZIP-B still localizes every head and reproduces the exact crowd count.
        As for the highly crowded scene (c), ZIP-B can still produce accurate density maps and counts.
    }
    \label{fig:more_vis}
\end{figure*}
\begin{figure*}[t]
    \centering
    \subfloat[Multi-scale scenes\label{fig:more_vis_scale}]{
        \includegraphics[width=\textwidth]{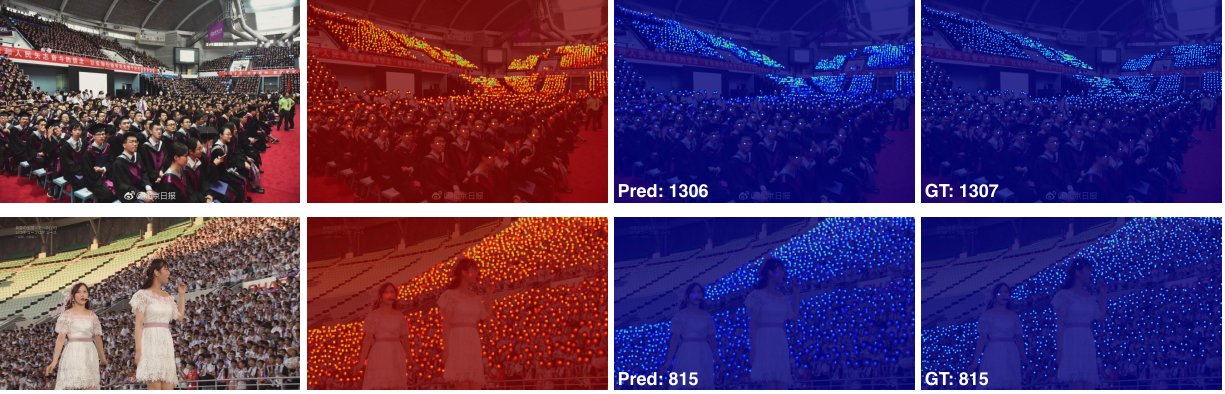}
    }
    \vfil
    \subfloat[Low-illumination examples\label{fig:more_vis_dim}]{
        \includegraphics[width=\textwidth]{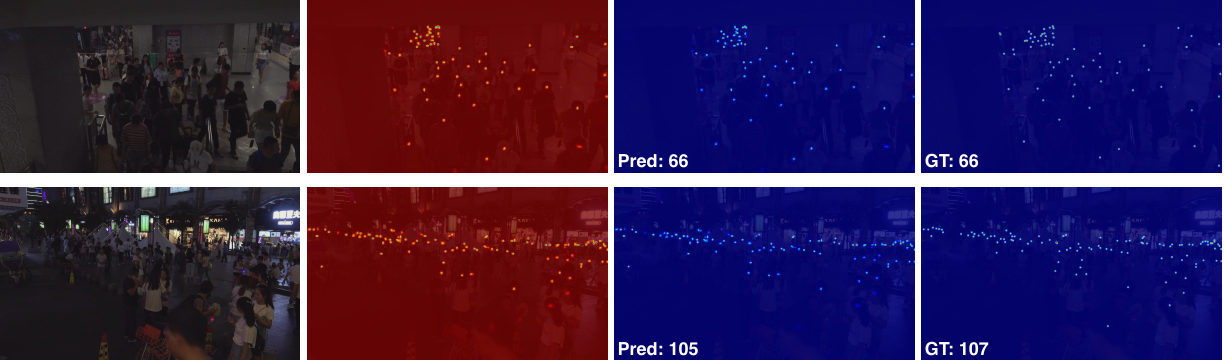}
    }
    \caption{
    Qualitative results on two challenging scenarios in NWPU-Crowd (predictions by ZIP-B): multi-scale (a) and low-illumination (b).
    For each row, the columns show (1) input image, (2) predicted structural-zero map, (3) predicted density map, and (4) ground-truth density map.
    In the multi-scale examples, the scenes contain a large variety of head sizes, ranging from large, close-up people to tiny, distant heads; ZIP-B suppresses background pixels while simultaneously detecting heads at all scales, yielding an accurate total count.
    In the low-illumination examples, the structural-zero branch can still filter the background, enabling the density branch to produce a clean, well-localized prediction despite poor lighting.}
    \label{fig:more_vis_challenging}
\end{figure*}

\subsection{Additional Qualitative Visualizations}

We curate seven images from the validation split of NWPU-Crowd to demonstrate the qualitative behavior of ZIP-B in five representative scenes. 
For clarity we organize the examples into a regular group (background-only, sparse, and crowded) and a challenging group (multi-scale and low-illumination). The results are shown in Fig.~\ref{fig:more_vis} and Fig.~\ref{fig:more_vis_challenging}, respectively.

\noindent\textbf{Background-only (Fig.~\ref{fig:more_vis_background}).} The zero-inflation branch assigns uniformly high zero-inflation probability throughout the image, while the predicted density map is effectively zero, matching the ground-truth count of 0.

\noindent\textbf{Sparse (Fig.~\ref{fig:more_vis_sparse}).} In a relatively sparse scene, the structural-zero map suppresses the background while isolating small blue blobs around true head centers; the resulting density maps reproduce the exact person counts.

\noindent\textbf{Crowded (Fig.~\ref{fig:more_vis_crowded}).} In a stadium-like scene containing several thousand people, the model still produces a clean, high-resolution density map. The total count deviates from ground truth by less than 0.5\%, confirming that ZIP modeling scales to extreme densities.

\noindent\textbf{Multi-scale (Fig.~\ref{fig:more_vis_scale}).} The foreground heads are large and the background heads extremely small. The structural-zero map adapts its granularity: it keeps broad foreground regions non-zero while aggressively filtering distant background pixels, enabling accurate detection across scales.

\noindent\textbf{Low-illumination (Fig.~\ref{fig:more_vis_dim}).} Despite poor lighting, the structural-zero branch reliably removes dark background regions, allowing the density branch to localize heads and deliver a near-exact total count.

Together, these qualitative results show that the explicit modeling of structural zeros enables ZIP-B to generalize from empty backgrounds to dim, highly congested, and multi-scale scenes without any segmentation supervision.

\begin{figure*}[!t]
    \centering
    \includegraphics[width=\textwidth]{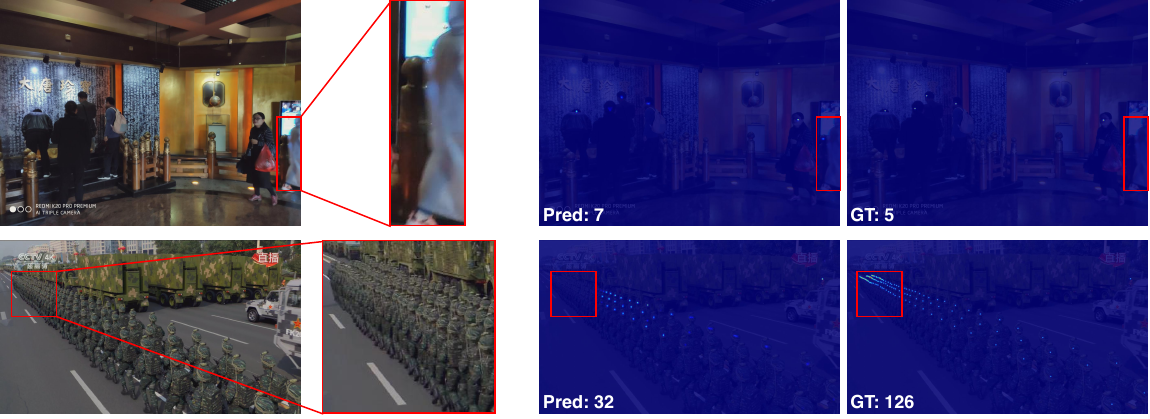}
    \caption{Two typical failure cases of ZIP-B. Left: input image with a zoom-in of the critical region (red box). Center: predicted density map with total count. Right: ground-truth density map with total count.}
    \label{fig:failures}
\end{figure*}
\section{Failure Analysis}

Despite its overall strong performance, ZIP-B still fails under certain visual ambiguities. Fig.~\ref{fig:failures} highlights two typical errors.
The top row corresponds to a false positive caused by a human-shaped artifact: in an indoor museum scene, a post along the rope barrier closely resembles a standing person in both shape and color, and as a result, ZIP-B misclassifies the post as a human.
In the bottom row, rows of soldiers wear camouflage that blends with the background trucks.
Although nearby soldiers are detected reliably, the model fails to pick up many small, low-contrast heads in the far field, leading to an under-count of -94.
These failure cases suggest that incorporating additional sensing modalities, particularly infrared imagery, may help suppress person-like artifacts and recover low-contrast targets, thereby reducing both false positives and false negatives. 
Exploring how the ZIP formulation can be extended to multi-modal crowd counting will be an important direction for our future work.


\section{Conclusion}
We presented ZIP, a zero-inflated Poisson framework for blockwise crowd counting. 
\minorrevreplace{By modeling structural-zero probability and positive-count rate separately,}{By jointly modeling a zero-inflation gate and a Poisson rate,} ZIP addresses the zero-heavy, discrete nature of point-derived count maps without requiring Gaussian-smoothed density targets or explicit segmentation supervision. 
Controlled ablations show that the ZIP likelihood improves over Poisson NLL, DMCount/EBC-style supervision, and direct-regression counterparts under matched backbones and training protocols. 
\minorrevreplace{Additional analyses validate the semantics and transfer behavior of the learned structural-zero map,}{Additional analyses show that the learned zero-inflation map is consistent with the proposed structural-zero interpretation and retains its ranking behavior across datasets,} and bin-sensitivity experiments show that the final bin-expectation implementation is robust to reasonable histogram-driven bin choices. 
Across four benchmarks and multiple model scales, ZIP achieves strong accuracy-efficiency trade-offs. 
Future work will extend this formulation to video crowd counting
and exploit temporal cues.

\bibliography{ref}



\clearpage
\appendices

\section{Proof of Lipschitz Continuity of ZIP NLL}
\label{app:proof_lip_zip}

\begin{proof}[Proof of Lemma~\ref{lem:1}]
We show that the per-block ZIP NLL in Eq.~\eqref{eqn:zip_block_nll} has bounded finite differences over $k\in\{0,1,\ldots,k_{\max}\}$.

\textbf{Case 1: $k=0$.}
For the transition from $0$ to $1$, we have
\begin{align}
    &\left|\ell_{\theta}^{\mathrm{ZIP}}(1)-\ell_{\theta}^{\mathrm{ZIP}}(0)\right| \notag\\
    &=
    \left|
    \log\left(\frac{\pi+(1-\pi)e^{-\lambda}}{1-\pi}\right)
    +\lambda-\log\lambda
    \right| \notag\\
    &\le
    \left|
    \log\left(\frac{\pi+(1-\pi)e^{-\lambda}}{1-\pi}\right)
    \right|
    +
    |\lambda-\log\lambda|.
    \label{eqn:zip_lip_case_zero}
\end{align}
Since
\begin{equation*}
    e^{-\lambda_{\max}}
    \le
    \pi+(1-\pi)e^{-\lambda}
    \le 1,
    \qquad
    1-\pi_{\max}
    \le
    1-\pi
    \le
    1-\pi_{\min},
\end{equation*}
we obtain
\begin{equation*}
    \left|
    \log\left(\frac{\pi+(1-\pi)e^{-\lambda}}{1-\pi}\right)
    \right|
    \le
    u_1,
\end{equation*}
where
\begin{equation*}
    u_1
    \coloneqq
    \max\left\{
    \left|\log\left(\frac{e^{-\lambda_{\max}}}{1-\pi_{\min}}\right)\right|,
    \left|\log\left(\frac{1}{1-\pi_{\max}}\right)\right|
    \right\}.
\end{equation*}
For $\lambda\ge1$, $\lambda-\log\lambda$ is monotonically increasing, and hence
\begin{equation*}
    |\lambda-\log\lambda|
    \le
    u_2,
    \qquad
    u_2\coloneqq \lambda_{\max}-\log\lambda_{\max}.
\end{equation*}
Therefore,
\begin{equation}
    \left|\ell_{\theta}^{\mathrm{ZIP}}(1)-\ell_{\theta}^{\mathrm{ZIP}}(0)\right|
    \le
    u_1+u_2.
    \label{eqn:zip_case_0_bound}
\end{equation}

\textbf{Case 2: $k\ge1$.}
For $k\in\{1,\ldots,k_{\max}-1\}$,
\begin{align*}
    \left|\ell_{\theta}^{\mathrm{ZIP}}(k+1)-\ell_{\theta}^{\mathrm{ZIP}}(k)\right|
    &=
    \left|\log\frac{k+1}{\lambda}\right| \\
    &\le
    \max\left\{
        \left|\log\frac{2}{\lambda_{\max}}\right|,
        \left|\log\frac{k_{\max}}{\lambda_{\min}}\right|
    \right\}.
\end{align*}
Let
\begin{equation*}
    u_3 \coloneqq
    \max\left\{
        \left|\log\frac{2}{\lambda_{\max}}\right|,
        \left|\log\frac{k_{\max}}{\lambda_{\min}}\right|
    \right\}.
\end{equation*}
Combining the two cases, define
\begin{equation*}
    l\coloneqq\max\{u_1+u_2,u_3\}.
\end{equation*}
Then, for all $k\in\{0,\ldots,k_{\max}-1\}$,
\begin{equation*}
    \left|\ell_{\theta}^{\mathrm{ZIP}}(k+1)-\ell_{\theta}^{\mathrm{ZIP}}(k)\right|
    \le l.
\end{equation*}
By telescoping, the adjacent-difference bound implies
\begin{equation*}
\left|\ell_{\theta}^{\mathrm{ZIP}}(k)-\ell_{\theta}^{\mathrm{ZIP}}(k')\right|
\le l|k-k'|
\end{equation*}
for any $k,k' \in \{0,\ldots,k_{\max}\}$.
Thus, $\ell_{\theta}^{\mathrm{ZIP}}(k)$ is $l$-Lipschitz continuous with respect to $k$ on the bounded discrete count domain.
\end{proof}

\section{Proof of Full-Image Lipschitz Continuity}
\label{app:proof_full_lip_zip}

\begin{proof}[Proof of Lemma~\ref{lem:2}]
Let $(\boldsymbol{z},\boldsymbol{\theta})$ and $(\boldsymbol{z}',\boldsymbol{\theta}')$ be two annotation-parameter pairs.
Using the definition of $\mathcal{L}_{\mathrm{ZIP}}$ in Eq.~\eqref{eqn:zip_image_nll},
\begin{align*}
    &\left|
    \mathcal{L}_{\mathrm{ZIP}}(\boldsymbol{z};\boldsymbol{\theta})
    -
    \mathcal{L}_{\mathrm{ZIP}}(\boldsymbol{z}';\boldsymbol{\theta}')
    \right| \\
    &\le
    \frac{1}{s}\sum_{i=1}^{s}
    \left|
    \ell_{\theta_i}^{\mathrm{ZIP}}(z_i)
    -
    \ell_{\theta_i}^{\mathrm{ZIP}}(z_i')
    \right|
    +
    \frac{1}{s}\sum_{i=1}^{s}
    \left|
    \ell_{\theta_i}^{\mathrm{ZIP}}(z_i')
    -
    \ell_{\theta_i'}^{\mathrm{ZIP}}(z_i')
    \right|.
\end{align*}
By Lemma~\ref{lem:1}, the first term is bounded by
\begin{equation*}
    \frac{l}{s}\|\boldsymbol{z}-\boldsymbol{z}'\|_1.
\end{equation*}
For the second term, compactness of $\Theta$ and differentiability of $\ell_{\theta}^{\mathrm{ZIP}}(k)$ with respect to $\theta$ imply
\begin{equation*}
    M\coloneqq
    \sup_{0\le k\le k_{\max},\theta\in\Theta}
    \|\nabla_{\theta}\ell_{\theta}^{\mathrm{ZIP}}(k)\|_1
    <\infty.
\end{equation*}
Thus,
\begin{equation*}
    \left|
    \ell_{\theta_i}^{\mathrm{ZIP}}(z_i')
    -
    \ell_{\theta_i'}^{\mathrm{ZIP}}(z_i')
    \right|
    \le
    M\|\theta_i-\theta_i'\|_1,
\end{equation*}
and therefore
\begin{equation*}
    \frac{1}{s}\sum_{i=1}^{s}
    \left|
    \ell_{\theta_i}^{\mathrm{ZIP}}(z_i')
    -
    \ell_{\theta_i'}^{\mathrm{ZIP}}(z_i')
    \right|
    \le
    \frac{M}{s}\|\boldsymbol{\theta}-\boldsymbol{\theta}'\|_1.
\end{equation*}
Combining both bounds gives
\begin{equation*}
    \left|
    \mathcal{L}_{\mathrm{ZIP}}(\boldsymbol{z};\boldsymbol{\theta})
    -
    \mathcal{L}_{\mathrm{ZIP}}(\boldsymbol{z}';\boldsymbol{\theta}')
    \right|
    \le
    \frac{l}{s}\|\boldsymbol{z}-\boldsymbol{z}'\|_1
    +
    \frac{M}{s}\|\boldsymbol{\theta}-\boldsymbol{\theta}'\|_1.
\end{equation*}
Hence $\mathcal{L}_{\mathrm{ZIP}}$ is $L_{\mathrm{ZIP}}$-Lipschitz with respect to the joint $\ell_1$ norm, with any
\begin{equation*}
    L_{\mathrm{ZIP}}\ge \frac{\max\{l,M\}}{s}.
\end{equation*}
\end{proof}

\begin{revblock}
\section{Poisson NLL Counterpart}
\label{app:poisson_bound}

For standard Poisson NLL,
\begin{equation*}
    \ell_{\lambda}^{\mathrm{P}}(k)=\lambda-k\log\lambda+\log k!,
    \qquad
    \lambda\in[\lambda_{\min},\lambda_{\max}],
\end{equation*}
we have, for adjacent counts,
\begin{equation*}
    \left|\ell_{\lambda}^{\mathrm{P}}(k+1)-\ell_{\lambda}^{\mathrm{P}}(k)\right|
    =
    \left|\log\frac{k+1}{\lambda}\right|,
\end{equation*}
which is bounded for $0\le k\le k_{\max} - 1$ and $\lambda\in[\lambda_{\min},\lambda_{\max}]$. 
The derivative with respect to $\lambda$ is
\begin{equation*}
    \left|\frac{\partial\ell_{\lambda}^{\mathrm{P}}(k)}{\partial\lambda}\right|
    =
    \left|1-\frac{k}{\lambda}\right|,
\end{equation*}
which is also bounded under the same assumptions. 
Therefore, the full-image Poisson NLL,
\begin{equation*}
    \mathcal{L}_{\mathrm{P}}(\boldsymbol z;\boldsymbol\lambda)
    =
    \frac{1}{s}\sum_{i=1}^{s}\ell_{\lambda_i}^{\mathrm{P}}(z_i),
\end{equation*}
is Lipschitz continuous with some finite constant $L_{\mathrm{P}}$. 
Applying the same Rademacher-complexity argument as in Theorem~\ref{thm:main} gives Eq.~\eqref{eqn:poisson_bound}.
\end{revblock}

\section{Extension to the Total Loss}
\label{app:total_loss_bound}

The same argument can be extended from ZIP NLL to the total training objective in Eq.~\eqref{eqn:total_loss}. 
Under the same bounded-logit assumption for the prediction heads,
the cross-entropy term is Lipschitz with a finite constant
$L_{\mathrm{CE}}$. Thus, its empirical Rademacher complexity is bounded by
$L_{\mathrm{CE}}s_+\mathfrak{R}_{\mathcal S}(\mathcal H)$, where $s_+$
denotes the number of blocks with nonzero counts and $0\le s_+\le s$.
Combining this with the bound for ZIP NLL and the counting-loss bound
in~\cite{wang2020distribution}, we obtain
\begin{align*}
    \mathcal{R}_{\mathcal D}(f_{\mathcal S}^{\mathrm{total}},
    \mathcal{L}_{\mathrm{total}})
    &\le
    \mathcal{R}_{\mathcal D}(f_{\mathcal D}^{\mathrm{total}},
    \mathcal{L}_{\mathrm{total}}) \\
    &\quad + 2(L_{\mathrm{ZIP}} + 1 + \omega L_{\mathrm{CE}})
    s \mathfrak{R}_{\mathcal S}(\mathcal H) \\
    &\quad + 5B\sqrt{\frac{2\log(8/\delta)}{K}},
\end{align*}
where $B=B_{\mathrm{ZIP}}+ \omega B_{\mathrm{CE}}+B_{\mathrm{count}}$ is a uniform upper bound on $\mathcal{L}_\mathrm{total}$.

%
%
%
%

\section{Data Augmentation}
\label{sec:data_augmentation}

\begin{table}[t]
\centering
\caption{
Dataset-specific training size and block size settings.
}
\label{tab:new_aug}
\begin{tabular}{l|cc}
\toprule
\textbf{Dataset} & \textbf{Training Size} & \textbf{Block Size} \\ \midrule
ShanghaiTech A   & 224                 & 16     \\
ShanghaiTech B   & 448                 & 16     \\
UCF-QNRF         & 672                 & 32     \\
NWPU-Crowd       & 672                 & 16     \\ \bottomrule
\end{tabular}
\end{table}

We apply a combination of data augmentation techniques during training, largely following the EBC framework \cite{ma2024clip}. The transformation pipeline includes the following operations:
\begin{itemize}
    \item \textbf{RandomResizedCrop}: For each image, we randomly select a cropping scale from a predefined range $[0.75, 2.0]$. A region of $\texttt{train\_size} \times \texttt{scale}$ is cropped and then resized back to $\texttt{train\_size}$.
    This operation effectively amplifies the large blockwise count values, which follow a long-tailed distribution.
    \item \textbf{RandomHorizontalFlip}: The image is flipped horizontally with a probability of 0.5.
    \item \textbf{ColorJitter}: To simulate diverse lighting and color conditions, we independently perturb the brightness, contrast, and saturation within specified ranges ($\pm 0.15$, $\pm 0.15$, $\pm 1.0$), while disabling hue adjustment (set to 0.0). For NWPU-Crowd, we increase the brightness perturbation range to $\pm 0.20$ to better cope with extreme low-light scenes. Torchvision is used in this augmentation.
    \item \textbf{PepperSaltNoise}: A small fraction of pixels are randomly replaced with either black (pepper, value 0) or white (salt, value 1), introducing local binary noise. The noise levels are controlled by two parameters: saltiness (ratio of salt pixels) and spiciness (ratio of pepper pixels). These two parameters are fixed to 0.001 across all datasets.
\end{itemize}
We use different values for training resolution and block sizes and the details are presented in Table~\ref{tab:new_aug}.

\begin{table}[t]
\centering
\begin{revblock}
\caption{
Ablation of training resolution and block size on UCF-QNRF using the same VGG19-based ZIP model.
Only the training resolution and block size are changed.
}
\label{tab:ucf_resolution_block_ablation}
\begin{tabular}{c|cc|cc}
\toprule
\multirow{2}{*}{\textbf{Training Res.}} 
& \multicolumn{2}{c|}{\textbf{Block Size 16}} 
& \multicolumn{2}{c}{\textbf{Block Size 32}} \\ 
\cmidrule{2-5}
& \textbf{MAE} & \textbf{RMSE} & \textbf{MAE} & \textbf{RMSE} \\
\midrule
448 & 78.5 & 130.2 & 75.7 & 127.1 \\
672 & 75.0 & 131.6 & \textbf{73.0} & \textbf{126.1} \\
896 & 77.9 & 133.6 & 75.9 & 130.3 \\
\bottomrule
\end{tabular}%
\end{revblock}
\end{table}

\begin{revblock}
\section{Training Resolution and Block Size}
\label{app:training_block_size}

Similar to existing density-map-based crowd counting methods, our ZIP framework also needs to determine the crop size used in training and the output downsampling rate of the prediction target. 
The training crop resolution affects memory cost, field-of-view coverage, and scale variation.
The downsampling rate (\ie the block size in ZIP) specifies the spatial granularity of the blockwise count map.
A smaller block size preserves finer spatial detail, but makes the supervision more zero-dominated and more sensitive to spatial annotation noise; a larger block size alleviates sparsity and stabilizes blockwise counts, but coarsens the spatial supervision.

As reported in Table~\ref{tab:new_aug}, we use training sizes of 224, 448, 672, and 672 for ShanghaiTech A, ShanghaiTech B, UCF-QNRF, and NWPU-Crowd, respectively.
We use block size 16 for ShanghaiTech A, ShanghaiTech B, and NWPU-Crowd, and block size 32 for UCF-QNRF.
\end{revblock}

\begin{revblock}
The UCF-QNRF setting is selected according to Table~\ref{tab:ucf_resolution_block_ablation}.
For all evaluated training resolutions, block size 32 gives lower MAE and RMSE than block size 16.
At training resolution 672, increasing the block size from 16 to 32 reduces the MAE from 75.0 to 73.0 and the RMSE from 131.6 to 126.1.
The best overall performance is achieved with training resolution 672 and block size 32, which is adopted for UCF-QNRF in the main experiments.

This result indicates that UCF-QNRF benefits from a coarser block grid, which produces more stable blockwise count targets while retaining sufficient spatial structure.
For NWPU-Crowd, the default block size of 16 is kept to preserve finer spatial localization.
These settings show that block size is not determined by image resolution alone, but also depends on local crowd density and the stability of the induced blockwise count targets.
\end{revblock}

\begin{figure*}[t]
    \centering
    \includegraphics[width=\textwidth]{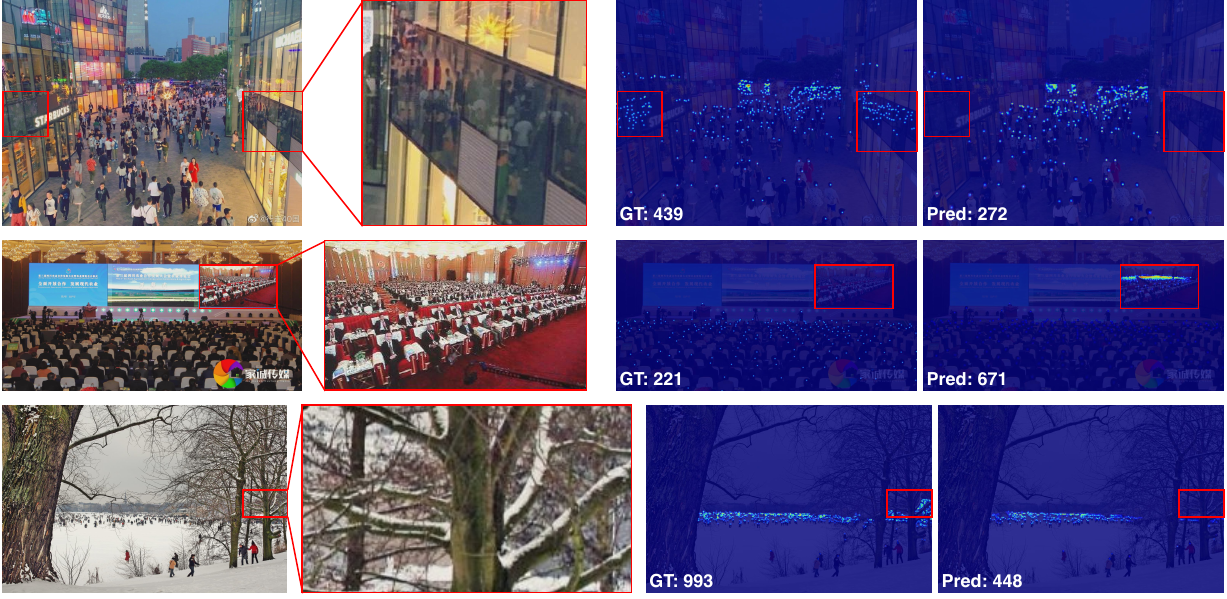}
    \caption{
    Examples of annotation noise in NWPU-Crowd.
    Left column: input images with zoom-in views of the regions outlined in red.
    Center column: ground-truth density map with total count. Right column: density map with total count predicted by ZIP-B.
    Top row: mirror reflections labeled as real people. Middle row: photo-realistic audience on a display screen left unlabeled. Bottom row: snow-covered tree branches mislabeled as a dense crowd.
    }
    \label{fig:annotation_noise}
\end{figure*}

\section{Detailed Bin Generation}
\label{app:bin_generation_details}

\begin{revblock}
This section provides the full bin-generation rules and the exact bin configurations used in the sensitivity study in Sec.~\ref{sec:bin_sensitivity}.
All bin boundaries and centers are computed from training-set blockwise count statistics only; validation and test annotations are not used.

\noindent\textbf{Frequency-aware default bins.}
For the default setting, we apply a shared frequency-aware rule to the training-set blockwise count histogram. 
Frequent low counts are kept as singleton bins, while rarer high counts are grouped into progressively wider intervals.
Specifically, counts 0--10 are assigned singleton bins when enough samples are available, counts 11--20 are grouped into two-valued bins, counts 21--29 into three-valued bins, and counts $\ge 30$ into four-valued bins.
We stop this pattern and use an open-ended tail bin $[t,\infty)$ once further splitting the tail would create bins with less than 0.001\% of all blockwise count values.
Each non-singleton bin center is the empirical mean of training-set blockwise counts that fall in that bin.
The resulting default configurations are:
\begin{itemize}
    \item \textbf{ShanghaiTech A (block size = 16):} bins $[0,0]$, $[1,1]$, $\ldots$, $[10,10]$, $[11,12]$, $[13,14]$, $[15,\infty)$; centers 0.0, 1.0, $\ldots$, 10.0, 11.38, 13.38, 16.26.
    \item \textbf{ShanghaiTech B (block size = 16):} bins $[0,0]$, $[1,1]$, $\ldots$, $[8,8]$, $[9,\infty)$; centers 0.0, 1.0, $\ldots$, 8.0, 10.16.
    \item \textbf{UCF-QNRF (block size = 32):} bins $[0,0]$, $[1,1]$, $\ldots$, $[10,10]$, $[11,12]$, $[13,14]$, $[15,16]$, $[17,18]$, $[19,20]$, $[21,23]$, $[24,26]$, $[27,29]$, $[30,33]$, $[34,\infty)$; centers 0.0, 1.0, $\ldots$, 10.0, 11.43, 13.43, 15.44, 17.44, 19.43, 21.83, 24.85, 27.87, 31.24, 38.86.
    \item \textbf{NWPU-Crowd (block size = 16):} bins $[0,0]$, $[1,1]$, $\ldots$, $[9,9]$, $[10,\infty)$; centers 0.0, 1.0, $\ldots$, 9.0, 12.16.
\end{itemize}

\noindent\textbf{Histogram-driven alternatives in Table~\ref{tab:bin_sensitivity}.}
The four alternative settings are also generated from training-set histograms.
Auto-Hist keeps the zero bin separate, preserves a positive-count singleton prefix when supported by the data, and then places tail cuts using positive-count cumulative distribution levels.
Coarse-Hist uses fewer cuts to test a lower-resolution discretization, while Fine-Hist uses a longer singleton prefix and finer tail splitting.
Shared-16 pools the ShanghaiTech A and B training histograms at block size 16 and uses one shared Auto-Hist configuration for both datasets.
The exact configurations used in Table~\ref{tab:bin_sensitivity} are:
\begin{itemize}
    \item \textbf{ShanghaiTech A, Auto-Hist:} bins $[0,0]$, $[1,1]$, $[2,2]$, $[3,3]$, $[4,4]$, $[5,5]$, $[6,6]$, $[7,7]$, $[8,8]$, $[9,9]$, $[10,10]$, $[11,11]$, $[12,14]$, $[15,\infty)$; centers 0.0, 1.0, 2.0, 3.0, 4.0, 5.0, 6.0, 7.0, 8.0, 9.0, 10.0, 11.0, 12.69, 16.22.
    \item \textbf{ShanghaiTech A, Coarse-Hist:} bins $[0,0]$, $[1,1]$, $[2,3]$, $[4,5]$, $[6,8]$, $[9,\infty)$; centers 0.0, 1.0, 2.22, 4.33, 6.65, 10.56.
    \item \textbf{ShanghaiTech A, Fine-Hist:} bins $[0,0]$, $[1,1]$, $[2,2]$, $[3,3]$, $[4,4]$, $[5,5]$, $[6,6]$, $[7,7]$, $[8,8]$, $[9,9]$, $[10,10]$, $[11,11]$, $[12,12]$, $[13,14]$, $[15,\infty)$; centers 0.0, 1.0, 2.0, 3.0, 4.0, 5.0, 6.0, 7.0, 8.0, 9.0, 10.0, 11.0, 12.0, 13.37, 16.22.
    \item \textbf{Shared-16 for ShanghaiTech A/B:} bins $[0,0]$, $[1,1]$, $[2,2]$, $[3,3]$, $[4,4]$, $[5,5]$, $[6,6]$, $[7,7]$, $[8,8]$, $[9,9]$, $[10,10]$, $[11,11]$, $[12,14]$, $[15,\infty)$; centers 0.0, 1.0, 2.0, 3.0, 4.0, 5.0, 6.0, 7.0, 8.0, 9.0, 10.0, 11.0, 12.70, 16.21.
    \item \textbf{ShanghaiTech B, Auto-Hist:} bins $[0,0]$, $[1,1]$, $[2,2]$, $[3,3]$, $[4,4]$, $[5,5]$, $[6,6]$, $[7,7]$, $[8,8]$, $[9,11]$, $[12,\infty)$; centers 0.0, 1.0, 2.0, 3.0, 4.0, 5.0, 6.0, 7.0, 8.0, 9.58, 12.84.
    \item \textbf{ShanghaiTech B, Coarse-Hist:} bins $[0,0]$, $[1,1]$, $[2,2]$, $[3,4]$, $[5,6]$, $[7,\infty)$; centers 0.0, 1.0, 2.0, 3.25, 5.30, 8.18.
    \item \textbf{ShanghaiTech B, Fine-Hist:} bins $[0,0]$, $[1,1]$, $[2,2]$, $[3,3]$, $[4,4]$, $[5,5]$, $[6,6]$, $[7,7]$, $[8,8]$, $[9,9]$, $[10,10]$, $[11,11]$, $[12,\infty)$; centers 0.0, 1.0, 2.0, 3.0, 4.0, 5.0, 6.0, 7.0, 8.0, 9.0, 10.0, 11.0, 12.84.
\end{itemize}
\end{revblock}

\section{Annotation Noise}

Although NWPU-Crowd is currently the largest high-resolution benchmark, we observed non-negligible label noise.
Fig.~\ref{fig:annotation_noise} presents three typical failure modes on the validation split of this dataset: over-labeled reflections (top row), under-labeled screen display (middle row), and complex natural textures being mislabeled as crowds (bottom row).

In the top row, the red box highlights strong reflections of pedestrians on a glass facade.
These mirror images have been annotated as real heads, yet they are visually less consistent with surrounding people, resulting a total count of 439.
ZIP-B correctly ignores the reflections, producing a lower but more realistic count of 272.

The middle row depicts a conference hall is shown on a large display.
The audience visible inside the screen looks photo-realistic, but none of these heads are annotated (GT = 221).
The model nevertheless detects them, yielding a higher prediction (671). 
Qualitatively, the additional detections correspond to the people shown on the screen.

In the bottom row, a snowy park scene is shown where the red box highlights complex textures on snow-covered tree branches.
These natural textures have been mistakenly annotated as a dense crowd in the background, leading to a grossly overestimated ground truth count of 993.
Our model is more robust to this type of noise, largely ignoring the mislabeled texture and predicting a more plausible count of 448.

These examples underline two points: (i) reported errors can stem from annotation artifacts rather than model failure, and (ii) the structural-zero + ZIP formulation is robust enough to down-weight unrealistic cues while still detecting plausible but unlabeled heads. 
Future benchmarks would benefit from more consistent annotation protocols, confidence tags to flag uncertain regions, and a multi-annotator verification process to improve label quality and consistency.

\begin{figure}[!t]
    \centering
    \subfloat[Block size = 8\label{fig:count_distribution_8}]{
        \includegraphics[width=\linewidth]{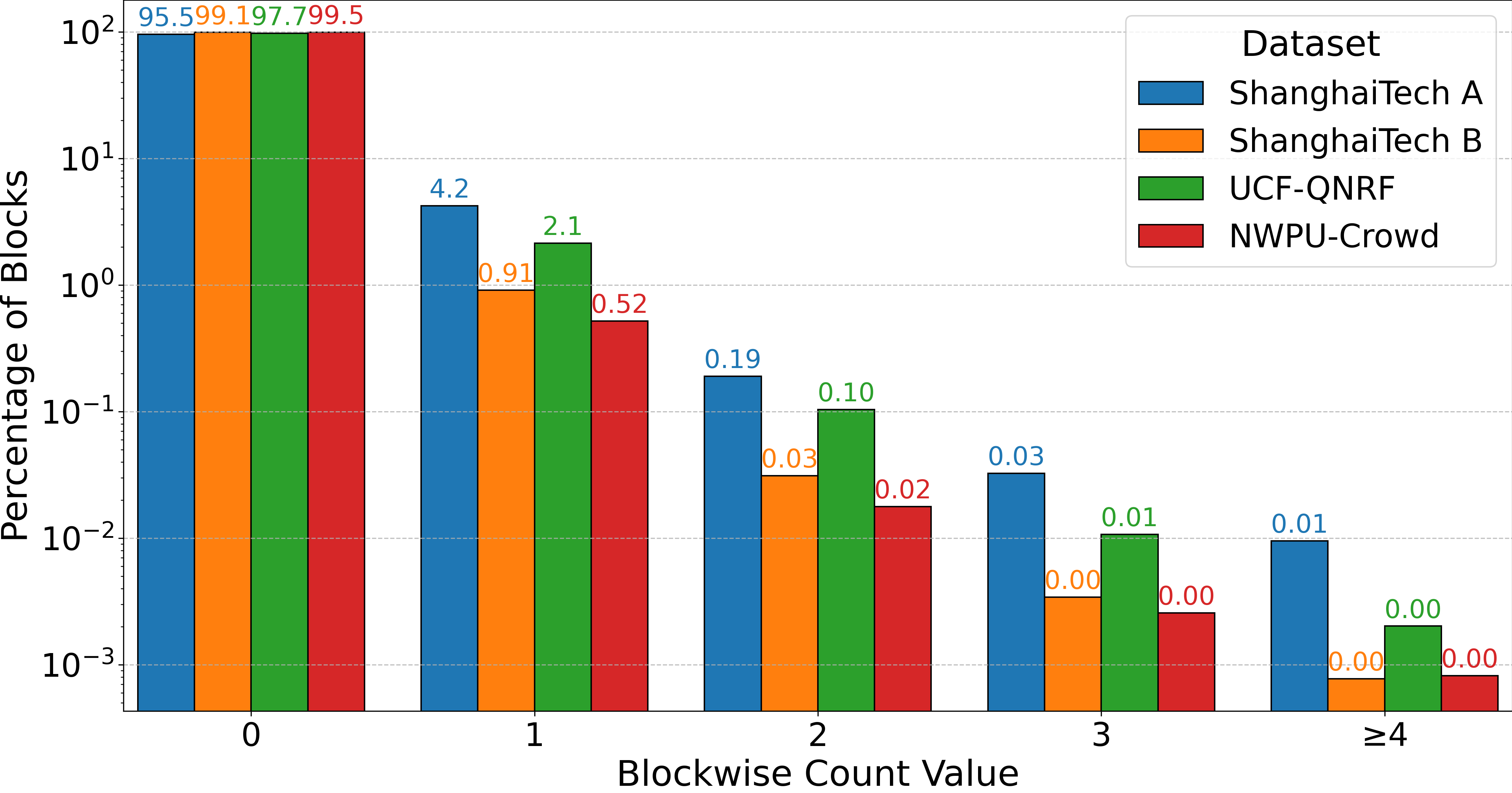}
    }
    \vfil
    \subfloat[Block size = 16\label{fig:count_distribution_16}]{
        \includegraphics[width=\linewidth]{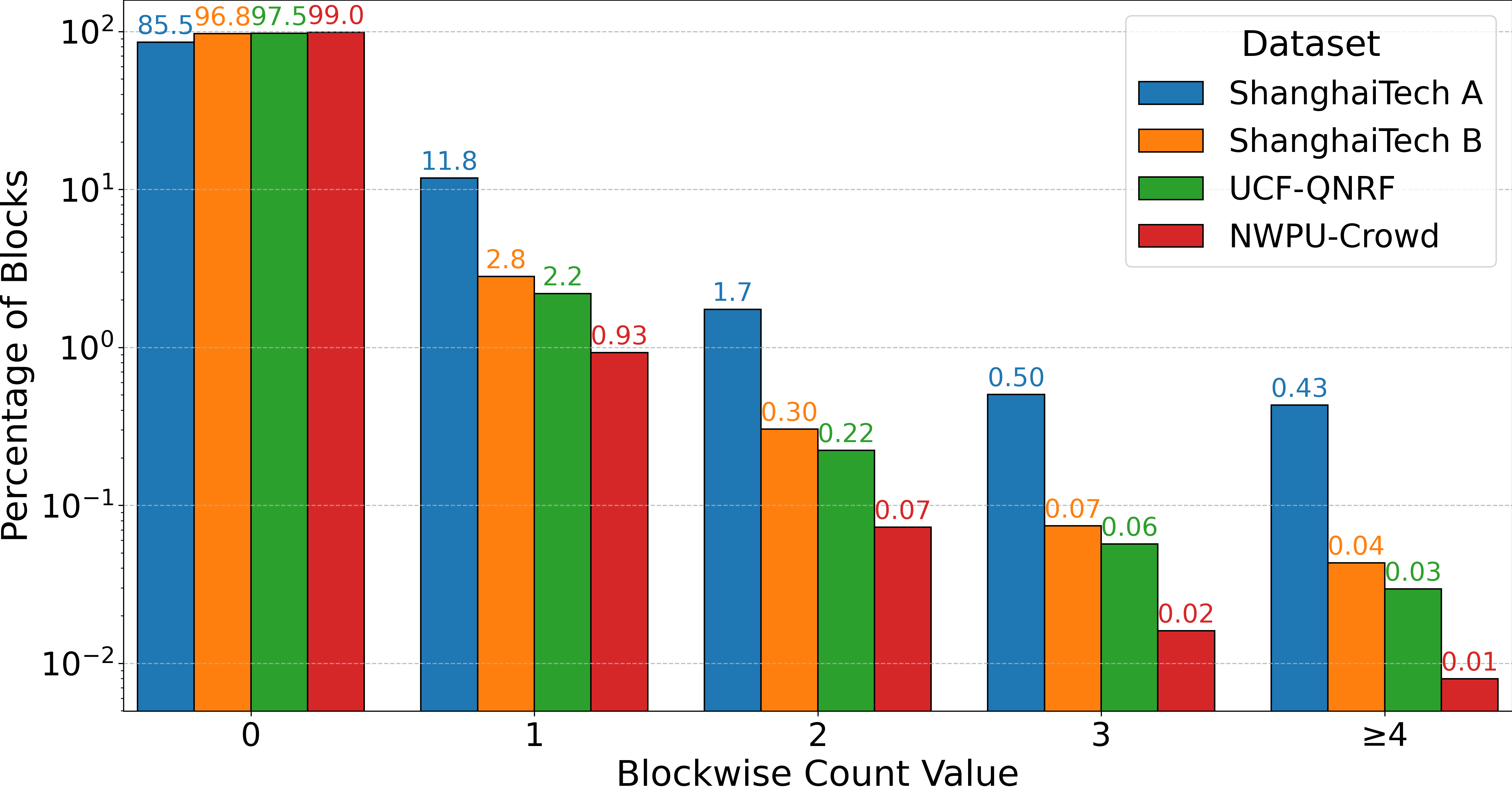}
    }
    \vfil
    \subfloat[Block size = 32\label{fig:count_distribution_32}]{
        \includegraphics[width=\linewidth]{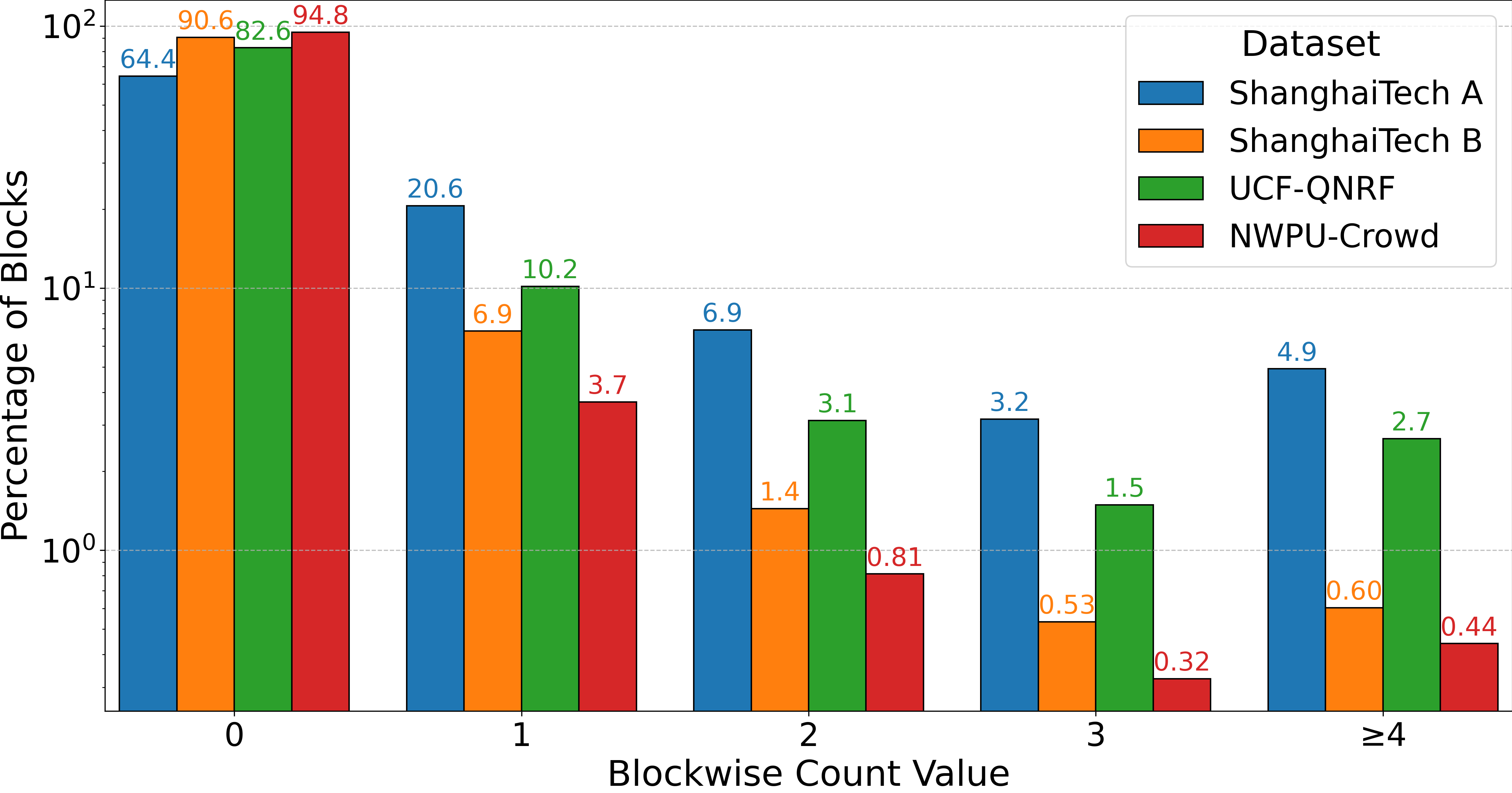}
    }
    \caption{Blockwise count distributions across common crowd counting benchmarks at different block sizes (8, 16, 32). For each dataset, we first generate ground-truth pixel-level density maps from point annotations, without using Gaussian smoothing. Then, we slide a $B{\times}B$ window with stride~1 across the map, summing the density within the window to obtain its count. Bars show the percentage of windows with count values $0,1,2,3,\ge4$ (log-scale $y$-axis).}
    \label{fig:count_distribution}
\end{figure}

\begin{table*}[!t]
\centering
\begin{minorrevblock}
\caption{Dataset-level excess-zero and dispersion analysis at the block sizes used in the main experiments. The fitted-Poisson zero probability is defined in Eq.~\eqref{eqn:fitted_poisson_zero}, the excess-zero gap $\Delta_0$ in Eq.~\eqref{eqn:excess_zero_gap}, and the empirical variance-to-mean ratio $D_{\mathrm{emp}}$ in Eq.~\eqref{eqn:empirical_dispersion}. Empirical zero percentages are those shown in Fig.~\ref{fig:count_distribution}.}
\label{tab:block_count_fit}
\setlength{\tabcolsep}{9pt}
\renewcommand{\arraystretch}{1.05}
\begin{tabular}{lccccc}
\toprule
\textbf{Dataset} &
$\boldsymbol{B}$ &
\textbf{Emp. Zero (\%)} &
\textbf{Pois. Zero (\%)} &
$\boldsymbol{\Delta_0}$ \textbf{(pp)} &
$\boldsymbol{D_{\mathrm{emp}}}$ \\
\midrule
ShanghaiTech A & 16 & 85.5 & 82.6 & +2.9  & 1.737 \\
ShanghaiTech B & 16 & 96.8 & 96.2 & +0.6  & 1.466 \\
UCF-QNRF       & 32 & 82.6 & 68.0 & +14.6 & 4.446 \\
NWPU-Crowd     & 16 & 99.0 & 97.8 & +1.2  & 1.652 \\
\bottomrule
\end{tabular}
\end{minorrevblock}
\end{table*}

\section{Blockwise Count Distribution}

For each dataset's \textbf{training split}, we form a pixel-level density map by placing 1 at each annotated head location and 0 elsewhere, without Gaussian smoothing being applied.
We then slide an axis-aligned $B{\times}B$ window over the map with stride~1 pixel and record the integer count equal to the sum of densities in that window (equivalently, the number of annotated heads whose pixel coordinates fall inside the window).
We repeat for $B \in \{8,16,32\}$ and aggregate counts over all windows from all images, converting to percentages.
Fig.~\ref{fig:count_distribution} illustrates the results.
\minorrev{
Under the block sizes used in the main experiments, the empirical proportions of zero-count blocks are 85.5\% for ShanghaiTech A ($B=16$), 96.8\% for ShanghaiTech B ($B=16$), 82.6\% for UCF-QNRF ($B=32$), and 99.0\% for NWPU-Crowd ($B=16$).
}
As the block size (window size) $B$ increases, zero percentages drop because each window integrates over a larger receptive field and is therefore more likely to include at least one head annotation.
\minorrevreplace{However, even at $B{=}32$ the distributions remain heavily zero-inflated. For instance, $>80\%$ zeros in UCF-QNRF and $>90\%$ in ShanghaiTech B and NWPU-Crowd. This indicates that annotation sparsity is a persistent regime rather than an artifact of using very fine windows.}
{Thus, the blockwise targets remain strongly zero-heavy at the block sizes used for training, rather than only at the standard $8\times8$ block size discussed in the motivation.}
Note that increasing $B$ also coarsens supervision: spatial detail about where heads occur \emph{within} a window is lost, making it harder for a crowd counting model to localize density peaks and learn fine-grained cues.
In the extreme limit, $B$ equal to the entire image collapses supervision to a single global count label; the learning problem degenerates to (image-level) weakly supervised counting with no spatial guidance at all.
Our ZIP formulation is motivated by operating in the practically useful regime of relatively small block size, where supervision is highly sparse yet still spatially informative, necessitating an explicit treatment of excess zeros.

\begin{minorrevblock}
\noindent\textbf{Poisson fit and excess-zero analysis.}
To determine whether the observed zero frequencies exceed those expected under a standard count model, we fit a separate Poisson distribution to each dataset's training-set blockwise counts at the block size used for training.
Let $Y$ denote a blockwise count and $y_1,\ldots,y_N$ its observed values for a given dataset.
The maximum-likelihood Poisson rate and its implied zero probability are
\begin{align}
    \hat{\lambda}
    &= \frac{1}{N}\sum_{i=1}^{N} y_i, \notag\\
    \hat{q}_{0}
    &= P(Y=0)
     = \exp\!\left(-\hat{\lambda}\right).
    \label{eqn:fitted_poisson_zero}
\end{align}
With $p_0$ denoting the empirical zero proportion, the excess-zero gap is
\begin{equation}
    \Delta_0
    = p_{0}
    - \hat{q}_{0}.
    \label{eqn:excess_zero_gap}
\end{equation}
A positive $\Delta_0$ means that the empirical distribution contains more zeros than expected under the fitted standard Poisson distribution.

We also evaluate the Poisson equidispersion assumption using the empirical variance-to-mean ratio
\begin{equation}
    D_{\mathrm{emp}}
    =
    \frac{\widehat{\mathrm{Var}}(Y)}
         {\widehat{\mathbb{E}}(Y)}.
    \label{eqn:empirical_dispersion}
\end{equation}
A standard Poisson distribution satisfies $\mathrm{Var}(Y)=\mathbb{E}(Y)$ and therefore has $D=1$.
Thus, $D_{\mathrm{emp}}>1$ indicates over-dispersion, meaning that the empirical blockwise counts vary more than a standard Poisson distribution can represent at the same mean.
\end{minorrevblock}

\begin{minorrevblock}
As shown in Table~\ref{tab:block_count_fit}, the empirical zero proportions exceed their fitted-Poisson expectations on all four datasets.
The corresponding excess-zero gaps are 2.9, 0.6, 14.6, and 1.2 percentage points for ShanghaiTech A, ShanghaiTech B, UCF-QNRF, and NWPU-Crowd, respectively, with the largest discrepancy observed on UCF-QNRF.
Moreover, $D_{\mathrm{emp}}$ ranges from 1.466 to 4.446 and exceeds one for every dataset, indicating overdispersion relative to the standard Poisson model.
Thus, the fitted Poisson distributions do not capture either the observed excess-zero mass or the empirical dispersion at the block sizes used for training.
\end{minorrevblock}

\begin{minorrevblock}
\noindent\textbf{Poisson--ZIP distribution-fit comparison.}
The ZIP distribution augments the Poisson model with a zero-inflation parameter, allowing additional probability mass at zero.
We fit a ZIP distribution to the same counts by maximum likelihood using Eq.~\eqref{eqn:zip_pmf}, and compare the Poisson and ZIP fits using total-variation distance over the full nonnegative-integer count support:
\begin{equation}
    \mathrm{TV}(p,p_M)
    = \frac{1}{2}\sum_{k=0}^{\infty}
       \left|p(k)-p_M(k)\right|, \, M\in\{\mathrm{P},\mathrm{ZIP}\}.
    \label{eqn:block_count_tv}
\end{equation}
A smaller TV distance indicates closer agreement with the full empirical blockwise count distribution.
\end{minorrevblock}

\begin{table}[!t]
\centering
\begin{minorrevblock}
\caption{Total-variation distance between the empirical blockwise count distribution and the fitted dataset-level Poisson and ZIP distributions, computed over the full nonnegative-integer count support; lower is better. The relative TV reduction is $100(\mathrm{TV}_{\mathrm{P}}-\mathrm{TV}_{\mathrm{ZIP}})/\mathrm{TV}_{\mathrm{P}}$; higher is better.}
\label{tab:block_count_tv}
\setlength{\tabcolsep}{4pt}
\begin{tabular}{lcccc}
\toprule
\textbf{Dataset} &
$\boldsymbol{B}$ &
\textbf{Poisson} &
\textbf{ZIP} &
\textbf{Rel. TV Red. (\%)} $\boldsymbol{\uparrow}$ \\
\midrule
ShanghaiTech A & 16 & 0.0395 & \textbf{0.0145} & 63.29 \\
ShanghaiTech B & 16 & 0.0089 & \textbf{0.0018} & 79.78 \\
UCF-QNRF       & 32 & 0.1798 & \textbf{0.0526} & 70.75 \\
NWPU-Crowd     & 16 & 0.0062 & \textbf{0.0014} & 77.42 \\
\bottomrule
\end{tabular}
\end{minorrevblock}
\end{table}

\begin{minorrevblock}
As shown in Table~\ref{tab:block_count_tv}, the fitted ZIP distribution yields a lower TV distance than the fitted Poisson distribution on all four datasets.
Relative to Poisson, ZIP reduces the TV distance by 63.29--79.78\%.
Because TV is evaluated over the full count support, these reductions reflect a closer fit to the overall empirical distribution, not only to its zero mass.
Together, the excess-zero, dispersion, and TV results support the proposed zero-inflated formulation.
\end{minorrevblock}

\end{document}